\documentclass[twoside]{article}

\usepackage[preprint]{aistats2026}
\usepackage[round]{natbib}

\usepackage[utf8]{inputenc} 
\usepackage[T1]{fontenc}    
\usepackage{hyperref}       
\usepackage{url}            
\usepackage{booktabs}       
\usepackage{amsfonts}       
\usepackage{nicefrac}       
\usepackage{microtype}      
\usepackage{xcolor}         

\usepackage{amsmath}
\usepackage{amssymb}
\usepackage{mathtools}
\usepackage{amsthm}
\usepackage{multirow}
\usepackage{caption}

\theoremstyle{plain}
\newtheorem{theorem}{Theorem}[section]
\newtheorem{proposition}[theorem]{Proposition}
\newtheorem{lemma}[theorem]{Lemma}

\theoremstyle{definition}

\theoremstyle{remark}
\newtheorem{remark}[theorem]{Remark}

\usepackage{bm}
\usepackage{lipsum}
\usepackage{dblfloatfix}
\usepackage{comment}
\usepackage{flushend}

\usepackage{xcolor}

\definecolor{revcolor}{HTML}{1565C0} 

\begin{document}

%

%

\runningauthor{Fernández-Hernández et al.}

\twocolumn[

\aistatstitle{Why $\beta_{1}=\beta_{2}$ Is Dynamically Special in {Adam}}

\aistatsauthor{
Alberto Fernández-Hernández
\And
Cristian Pérez-Corral
\And
Jose I. Mestre
}

\aistatsaddress{
Universitat Politècnica de València\\
Valencia, Spain\\
\texttt{a.fernandez@upv.es}
\And
Universitat Politècnica de València\\
Valencia, Spain\\
\texttt{cpercor@upv.es}
\And
Universitat Politècnica de València\\
Valencia, Spain\\
\texttt{jimesmir@upv.es}
}

\aistatsauthor{Manuel F. Dolz\And Enrique S. Quintana-Ortí}

\aistatsaddress{Universitat Jaume I\\ Castelló de la Plana, Spain\\ \texttt{dolzm@uji.es} \And Universitat Politècnica de València\\ Valencia, Spain\\ \texttt{quintana@disca.upv.es} }
]

\begin{abstract}
Adam has been at the core of large-scale training for almost a decade, yet the role of its two momentum parameters remains poorly understood. Recent work shows that tying $\beta_{1}=\beta_{2}$ can preserve Adam's strong performance despite collapsing two memory scales into one, raising a basic question: what becomes dynamically special when the memories are tied? We identify a concrete mechanism. In the continuous-time limit, each normalized-update coordinate decomposes into a sign component, an explicit magnitude-lag term proportional to the difference between the two memory times, and additional transition, curvature, and nonlinear ratio terms. This lag channel vanishes exactly when $\beta_{1}=\beta_{2}$, making the diagonal the unique regime in which this mismatch-induced response is structurally absent. A full-history discrete decomposition on real training gradients recovers this change in composition: tied updates are sign-dominated, whereas the lag term becomes substantial off the diagonal and leaves a comparatively small residual. Across six vision and language tasks, tied configurations also typically exhibit smoother update-norm trajectories. Overall, our results identify memory-scale mismatch as a concrete source of magnitude sensitivity in Adam and provide a mechanistic account of why tied momentum is dynamically distinctive.
\end{abstract}

\section{Introduction}

Adam \citep{Adam} has played a central role in neural network training for over a decade. Its combination of first-moment momentum \citep{Momentum} and an exponential moving average of squared gradients has made it a default choice across architectures and domains, from convolutional networks to LLMs. The original paper proposed the now standard defaults $\beta_{1}=0.9$ and $\beta_{2}=0.999$, and although there have been variations and some task-specific recommendations, such as using $\beta_{2}=0.95$ in certain LLMs \citep{Llama}, the rationale for keeping two separate momentum parameters and for how they should be related has remained largely unclear.

Recently, \citet{AdamSecretSauce} showed that tying $\beta_{1}=\beta_{2}$ can preserve Adam's strong performance. This is structurally surprising: Adam is built from two distinct exponential memories, yet collapsing their time scales can remain competitive. The question is therefore not only which hyperparameters work well, but what mechanism makes the diagonal $\beta_{1}=\beta_{2}$ dynamically special.

Gradient magnitudes mix useful descent information with problem scaling, anisotropy, parameterization, and stochastic variability. Modern optimizers increasingly control which parts of this magnitude information reach the update: sign methods remove coordinate-wise magnitudes, while normalization and matrix-based methods suppress them at coarser structural levels. Adam is subtler because magnitude enters through two history-dependent filters. When those filters track a changing signal at different rates, their mismatch can itself generate a magnitude-dependent response. This observation is the starting point of our analysis.

We show that this mismatch appears explicitly in Adam's normalized update. In continuous time, each coordinate contains a magnitude-lag term proportional to $(\tau_{2}-\tau_{1})\,d\log|g|/dt$, in addition to sign, transition, curvature, and nonlinear ratio effects. Hence $\beta_{1}=\beta_{2}$ is not merely one point in a favorable hyperparameter region: it is the unique regime in which this mismatch-induced channel is structurally absent for every admissible sign-stable trajectory. We then test whether this mechanism is realized in practice. A literal full-history decomposition of discrete Adam on training gradients finds a sign-dominated update on the diagonal and a substantial lag contribution off it, while full training across vision and language tasks shows a recurring smoother-update signature when the memories are tied.

Our contributions are as follows:
\begin{itemize}
\item We identify a mismatch-induced magnitude-lag channel in Adam's normalized update, proportional to $(\tau_{2}-\tau_{1})\,d\log|g|/dt$. Tying the memories removes this channel exactly while retaining the remaining history-dependent effects.

\item We show that this is not only a formal cancellation. A literal discrete
decomposition on independently trained real-gradient trajectories is
sign-dominated on the diagonal and assigns substantial weight to the lag term
off it, with a much smaller residual. Across six vision and language
model--dataset pairs, tied configurations also typically exhibit smoother
update-norm trajectories.

\end{itemize}

\section{Related Work}

Continuous-time analyses provide a natural framework for studying the memory effects of adaptive optimization. \citet{DaSilvaGazeau} develop a general ODE formulation encompassing Adam and several other first-order methods, and establish properties of the resulting flows. More closely related to our setting, \citet{MaWuE} investigate the dynamics of RMSProp and Adam, distinguishing a sign-gradient limit obtained with fixed momentum parameters from a memory-preserving limit in which the momentum parameters approach one as the step size vanishes. They identify different late-training regimes, including oscillations, spikes, and divergence, and observe that bringing $\beta_{1}$ and $\beta_{2}$ closer can lead to smoother convergence. These results establish both the relevance of continuous-time models and the importance of the relationship between Adam's momentum parameters. Our focus is different: rather than characterizing the resulting optimization regimes, we examine how the relative response times of the two moment filters enter the normalized update itself.

A complementary line of work seeks to understand the information retained by Adam's update. \citet{BallesHennig} interpret Adam as combining a sign component with magnitude modulation determined by relative gradient variance, separating these two effects to study their respective roles. Related evidence shows that sign descent with momentum can closely reproduce Adam's behavior in certain low-noise transformer settings \citep{KunstnerSignDescent}. These connections motivate taking the sign as a reference component, but leave a distinct dynamical question: even when a gradient coordinate retains its sign, its magnitude changes over time, and Adam's two exponential filters need not respond to those changes at the same rate.

The closest empirical and theoretical motivation for our work comes from \citet{AdamSecretSauce}. Through extensive language-model experiments, they show that tying $\beta_{1}=\beta_{2}$ preserves near-optimal performance across a range of training configurations, whereas carefully tuned signed-momentum methods retain a performance gap. Their theoretical contribution goes beyond this empirical observation. For equal momentum parameters, they derive an exact representation of Adam's update in terms of an online mean estimate and an associated variance estimate, and obtain these estimators from a Gaussian variational-inference formulation. They further establish that equal betas are necessary for their particular variance-estimation representation. This provides a statistical characterization of the tied regime and clarifies how it differs from pure signed momentum. Our question concerns a different property of the same regime: how unequal memory times introduce a response to changing gradient magnitudes through the interaction of the numerator and denominator. The necessity of equal betas for the variance representation does not characterize this temporal response, which requires examining the update along evolving gradient trajectories.

We investigate this interaction by separating the normalized update into a sign component, an explicit magnitude-lag contribution proportional to $(\tau_{2}-\tau_{1})\,d\log|g|/dt$, and remaining transition, curvature, and nonlinear ratio effects. The diagonal is distinguished by the exact cancellation of the explicit mismatch-induced term, not by complete invariance to gradient magnitudes or the disappearance of all history-dependent effects. We further examine whether this decomposition is informative beyond the continuous-time analysis: a full-history discrete counterpart is evaluated on gradients from independently trained models, and update-norm trajectories are studied across vision and language tasks. These diagnostics address the composition and temporal behavior of the update rather than establishing a general relationship between tied momentum and predictive performance.

More broadly, optimizers such as signSGD \citep{BernsteinSignSGD}, Lion \citep{Lion}, Shampoo and SOAP \citep{Shampoo,SOAP}, Muon \citep{Muon}, Scion \citep{Scion}, and RMNP \citep{deng2026rmnp} illustrate different ways of controlling the influence of gradient magnitudes at coordinate, matrix, or neuron level. We use this broader perspective to motivate the notion of structured magnitude sensitivity introduced in Section~3.1, where these mechanisms are discussed in more detail. Our analysis, however, concerns the specific temporal interaction between Adam's two memories rather than the design of a different normalization rule.

\section{Memory Mismatch and Magnitude Lag in Adam}
\label{sec:theory}

This section develops the main theoretical result: a mismatch between Adam's two memory times creates an explicit magnitude-lag channel in the normalized update. We first formalize \emph{structured sensitivity to gradient magnitudes}, then derive this channel and show that it vanishes exactly on the diagonal $\beta_{1}=\beta_{2}$.

\subsection{Structured Sensitivity to Gradient Magnitudes}
\label{subsec:structured-sensitivity}

A gradient-based optimizer generates a sequence of parameters
$(\bm{\theta}_{k})_{k \geq 0} \subset \mathbb{R}^{d}$ from the gradients
\[
\mathbf{g}_{k} := \nabla f(\bm{\theta}_{k}) \in \mathbb{R}^{d}
\]
through updates of the form
\[
\bm{\theta}_{k+1}
=
\bm{\theta}_{k}
-
\eta_{k}\mathbf{R}_{k},
\]
where $\eta_k$ is the learning rate and $\mathbf{R}_{k}$ is determined by the current gradient together with the internal state accumulated from previous gradients. Constructing $\mathbf{R}_{k}$ therefore requires deciding how the sign, magnitude, history, and structural organization of the gradient should influence the update.

This distinction matters because gradient magnitudes carry several effects simultaneously. They describe the local first-order variation of the objective, but also reflect curvature anisotropy, parameterization, stochastic variability, and the units in which different parameter blocks are represented. For example, consider
\[
f_{\kappa}(x,y)
=
\frac{1}{2}\left(x^{2}+\kappa y^{2}\right),
\qquad
\nabla f_{\kappa}(x,y)
=
\left(x,\kappa y\right),
\]
with $\kappa \gg 1$. At $(x,y)=(1,1)$, the second gradient coordinate is $\kappa$ times larger than the first. A raw gradient step transfers this factor directly to the update, although the imbalance is induced by the geometry of the objective rather than by a $\kappa$-fold difference in distance to the minimizer. The gradient remains the correct local differential; the optimizer must decide how literally its coordinate magnitudes should determine the available update.

Different geometries lead to different answers. Under a coordinate-wise update budget,
\[
\min_{\|\mathbf{u}\|_{\infty}\leq \rho}
\left\langle \mathbf{g},\mathbf{u}\right\rangle,
\]
the steepest first-order update is 
\(
\mathbf{u}^{\star}
=
-\rho\,\operatorname{sign}(\mathbf{g}).
\)
Thus, a sign update is not merely a crude approximation to the gradient. It is the exact steepest-descent direction for an $\ell_{\infty}$ trust region: it preserves the descent orthant while assigning the same budget to every active coordinate. Likewise, normalization under an $\ell_{2}$ budget removes one global magnitude, block normalization assigns separate budgets to parameter groups, and matrix normalization can suppress magnitudes associated with particular spectral directions. Each choice retains some gradient information and attenuates other information.

We use \emph{structured sensitivity to gradient magnitudes} to describe this design choice. The term ``structured'' refers to the units at which magnitudes are transformed. These units may be the whole gradient, individual coordinates, parameter groups, rows or columns of a weight matrix, or singular directions. The term ``sensitivity'' refers to how the resulting update changes when the corresponding magnitudes are changed while the retained structure is kept fixed.

More precisely, let $\mathcal{G}=\{G_{1},\ldots,G_{q}\}$ be a partition of the gradient coordinates and let
$\bm{\lambda}=(\lambda_{1},\ldots,\lambda_{q})\in\mathbb{R}_{>0}^{q}$.
Define the group-wise rescaling
\[
\bigl(T_{\bm{\lambda}}^{\mathcal{G}}\mathbf{g}\bigr)_{G_j}
=
\lambda_j\mathbf{g}_{G_j},
\qquad
j=1,\ldots,q.
\]
For a fixed optimizer state $\mathbf{s}_{k}$, the response of the update rule to this transformation is
\[
\bm{\Delta}_{\mathcal{G}}\mathbf{R}_{k}(\bm{\lambda})
:=
\mathbf{R}_{k}
\bigl(T_{\bm{\lambda}}^{\mathcal{G}}\mathbf{g}_{k};\mathbf{s}_{k}\bigr)
-
\mathbf{R}_{k}
\bigl(\mathbf{g}_{k};\mathbf{s}_{k}\bigr).
\]
Exact invariance is the limiting case
$\bm{\Delta}_{\mathcal{G}}\mathbf{R}_{k}(\bm{\lambda})=\mathbf{0}$.
More generally, the size and form of this response describe which magnitude perturbations reach the update and through which channels. Local sensitivity can be measured by differentiating this response with respect to $\log\lambda_j$ at $\bm{\lambda}=\mathbf{1}$.

This lens places several familiar optimizers on a common map. Gradient descent preserves all coordinate magnitudes and responds linearly to every rescaling. Normalized gradient descent removes a single global magnitude. signSGD \citep{BernsteinSignSGD} is invariant to arbitrary positive coordinate-wise rescalings, while RPROP \citep{RPROP} combines the same coordinate-wise sign information with step sizes determined by sign agreement over time. Lion \citep{Lion} applies a coordinate-wise sign transformation to a momentum quantity. For a matrix gradient
$\mathbf{G}=\mathbf{U}\bm{\Sigma}\mathbf{V}^{\mathsf{T}}$, the polar factor
$\mathbf{U}\mathbf{V}^{\mathsf{T}}$ preserves the singular directions while discarding the positive singular values; Muon \citep{Muon} approximates this type of orthogonalized matrix update. RMNP \citep{deng2026rmnp} replaces Muon's full-matrix preconditioner with row-wise momentum normalization, making each neuron-wise update invariant to a common positive rescaling of its corresponding momentum row. Scion \citep{Scion} constructs norm-constrained updates adapted to the structure of the parameter tensor. These mechanisms operate at different granularities, but all specify which gradient magnitudes are allowed to control the update.

Adam occupies an intermediate, history-dependent position. Its normalized update combines a first-moment memory with the square root of a second-moment memory. Consequently, coordinate-wise magnitudes are neither preserved directly nor removed exactly: they enter through the relative dynamics of two filters. A change in the current magnitude can affect the numerator and denominator at different rates, producing a lag between the two memories. Our analysis isolates this dynamic channel. The decomposition below isolates an explicit contribution proportional to
the difference between the two memory times and to the coordinate-wise
logarithmic drift of the gradient magnitude. Thus, the diagonal $\beta_{1}=\beta_{2}$ is the unique locus where the mismatch-induced magnitude-lag channel is structurally eliminated, while Adam's remaining history-dependent dynamics are retained.

This is the sense in which tied momentum changes Adam's structured sensitivity to gradient magnitudes. The claim concerns the internal composition of Adam's dynamics: the regime $\beta_{1}=\beta_{2}$ removes one specific route by which time-varying coordinate-wise magnitudes enter the normalized update. The resulting decomposition can then be measured directly along training trajectories.

\subsection{The Adam Flow}
\label{subsec:adam-flow}

Adam maintains two auxiliary sequences
$(\mathbf{m}_{k})_{k \geq 0}$ and
$(\mathbf{v}_{k})_{k \geq 0}$ that track first- and second-moment information:
\begin{align}
\mathbf{m}_{k+1}
&=
\beta_{1}\mathbf{m}_{k}
+
(1-\beta_{1})\mathbf{g}_{k},
\label{eq:adam-m}
\\
\mathbf{v}_{k+1}
&=
\beta_{2}\mathbf{v}_{k}
+
(1-\beta_{2})\mathbf{g}_{k}^{2},
\label{eq:adam-v}
\end{align}
where $\beta_{1},\beta_{2}\in(0,1)$ and
$\mathbf{g}_{k}^{2}$ denotes the coordinate-wise square of
$\mathbf{g}_{k}$. The parameter update is
\begin{equation}
\bm{\theta}_{k+1}
=
\bm{\theta}_{k}
-
\eta_k
\frac{\mathbf{m}_{k+1}}
{\sqrt{\mathbf{v}_{k+1}}+\varepsilon},
\label{eq:adam-theta}
\end{equation}
where $\varepsilon>0$ is a stabilization constant and all vector operations are coordinate-wise.

In practice, Adam commonly uses the bias-corrected moments
$\mathbf{m}_{k+1}/(1-\beta_{1}^{k+1})$ and
$\mathbf{v}_{k+1}/(1-\beta_{2}^{k+1})$. To expose the interaction between the two memories, the main derivation considers the raw recurrences
\eqref{eq:adam-m}--\eqref{eq:adam-theta} with $\varepsilon=0$.
Remarks~\ref{rem:bias-correction} and~\ref{rem:epsilon} in
Appendix~\ref{appendix:technical-details} describe the corresponding bias-corrected and positive-$\varepsilon$ dynamics.

Following continuous-time analyses of adaptive optimization methods
\citep{DaSilvaGazeau,MaWuE}, we interpret Adam as a time discretization of an underlying dynamical system. This formulation makes it possible to separate the coordinate-wise channels through which a changing gradient magnitude enters the normalized update and to identify the channel governed by the relation between $\beta_{1}$ and $\beta_{2}$.

Let $\Delta t>0$, set $t_k=k\Delta t$, and write $ \bm{\theta}_{k}\approx\bm{\theta}(t_k),
\mathbf{m}_{k}\approx\mathbf{m}(t_k),
\mathbf{v}_{k}\approx\mathbf{v}(t_k),$ and $
\mathbf{g}_{k}\approx\mathbf{g}(t_k)
=
\nabla f(\bm{\theta}(t_k)). $ Parameterize the discrete momentum coefficients as
\(
\beta_{1}
=
e^{-\Delta t/\tau_{1}},
\beta_{2}
=
e^{-\Delta t/\tau_{2}},
\)
with relaxation times $\tau_{1},\tau_{2}>0$, and scale the learning rate as
$\eta=\bar{\eta}\Delta t$ for $\bar{\eta}>0$. Formally passing to the limit
$\Delta t\to0$ in
\eqref{eq:adam-m}--\eqref{eq:adam-theta} yields
\begin{equation}
\label{eq:adam-flow}
\begin{aligned}
\tau_{1}\mathbf{m}'(t)
&=
-\mathbf{m}(t)+\mathbf{g}(t),
\\
\tau_{2}\mathbf{v}'(t)
&=
-\mathbf{v}(t)+\mathbf{g}(t)^{2},
\\
\bm{\theta}'(t)
&=
-\bar{\eta}
\frac{\mathbf{m}(t)}{\sqrt{\mathbf{v}(t)}},
\end{aligned}
\end{equation}
where
$\mathbf{g}(t)=\nabla f(\bm{\theta}(t))$ and all operations are coordinate-wise.

For each coordinate, define the normalized update
\begin{equation}
\label{eq:R-def}
\mathbf{R}(t)
:=
\frac{\mathbf{m}(t)}{\sqrt{\mathbf{v}(t)}},
\end{equation}
so that
$\bm{\theta}'(t)=-\bar{\eta}\mathbf{R}(t)$.
Appendix~\ref{appendix:continuum} provides the derivation of
\eqref{eq:adam-flow} and the relation between the relaxation times
$\tau_{1},\tau_{2}$ and the discrete coefficients
$\beta_{1},\beta_{2}$.

\subsection{Coordinate-wise Magnitude Dynamics in the Adam Flow}
\label{subsec:adam-gradient-dependence}

The discussion of Subsection \ref{subsec:structured-sensitivity}
compared optimizers through their response to structured changes in gradient magnitude at a fixed internal state. We now study the complementary dynamical question for Adam: how time-varying coordinate-wise magnitudes propagate through its two memories along a gradient trajectory. The key variable is the relative rate at which each magnitude changes.

\paragraph{Logarithmic magnitude drift.}
Let $I=[t_{0},t_{1}]$ be an interval on which every coordinate of
$\mathbf{g}(t)$ is nonzero. Since $\mathbf{g}$ is continuous, its coordinate-wise sign is constant on $I$, and we may write
\[
\mathbf{g}(t)
=
\operatorname{sign}\bigl(\mathbf{g}(t)\bigr)
\odot
|\mathbf{g}(t)|.
\]
We define the \emph{coordinate-wise logarithmic magnitude drift} by
\begin{equation}
\label{eq:delta-def}
\bm{\delta}(t)
:=
\frac{d}{dt}\log|\mathbf{g}(t)|
=
\frac{\mathbf{g}'(t)}{\mathbf{g}(t)},
\end{equation}
where the logarithm, quotient, and absolute value are taken coordinate-wise. Thus, $\delta_i(t)$ measures the instantaneous relative change of $|g_i(t)|$, rather than its additive change.

This representation separates baseline magnitude from magnitude variation. If
$\widetilde{\mathbf{g}}(t)=\mathbf{c}\odot\mathbf{g}(t)$ for a fixed
$\mathbf{c}\in\mathbb{R}_{>0}^{d}$, then
$\widetilde{\bm{\delta}}(t)=\bm{\delta}(t)$. If instead
$\widetilde{\mathbf{g}}(t)=\bm{\lambda}(t)\odot\mathbf{g}(t)$ for a positive time-dependent rescaling, then
\[
\widetilde{\bm{\delta}}(t)
=
\bm{\delta}(t)
+
\frac{\bm{\lambda}'(t)}{\bm{\lambda}(t)}.
\]
Accordingly, $\bm{\delta}(t)$ is insensitive to fixed coordinate units and records time-varying multiplicative changes. Over a short interval on which $\delta_i(t)$ is nearly constant,
\(
|g_i(t+\Delta t)|
\approx
|g_i(t)|\exp\bigl(\delta_i(t)\Delta t\bigr).
\)
Its discrete analogue is therefore the logarithmic ratio
\[
\log|g_{i,k+1}|-\log|g_{i,k}|
=
\log\left(\frac{|g_{i,k+1}|}{|g_{i,k}|}\right)
\]
on steps without a sign change.

The next proposition uses this variable to separate the Adam moments into an instantaneous tracking component, an explicit response to magnitude drift, a transition term determined by the initial moment state, and a curvature term controlled by the variation of the drift.

\begin{proposition}
\label{prop:m-v-delta}
Let $I=[t_{0},t_{1}]$ be a compact interval and let
$\mathbf{g}:I\to\mathbb{R}^{d}$ be a $C^{2}$ mapping such that
$g_i(t)\neq 0$ for every $t\in I$ and every $i\in\{1,\ldots,d\}$.
Let $\bm{\delta}(t)$ be defined by \eqref{eq:delta-def}, and set
\[
\Lambda
:=
\sup_{t\in I}\|\bm{\delta}(t)\|_{\infty},
\qquad
\Lambda'
:=
\sup_{t\in I}\|\bm{\delta}'(t)\|_{\infty}.
\]
Let $\mathbf{m}(t)$ and $\mathbf{v}(t)$ solve the first two equations of the Adam flow. Then, for every $t\in I$,
\begin{equation}
\label{eq:moment-drift-decomposition}
\begin{aligned}
\mathbf{m}(t)
&=
\mathbf{g}(t)
\odot
\bigl(\mathbf{1}-\tau_{1}\bm{\delta}(t)\bigr)
+
\mathbf{T}_{m}(t)
+
\mathbf{C}_{m}(t),
\\
\mathbf{v}(t)
&=
\mathbf{g}(t)^{2}
\odot
\bigl(\mathbf{1}-2\tau_{2}\bm{\delta}(t)\bigr)
+
\mathbf{T}_{v}(t)
+
\mathbf{C}_{v}(t).
\end{aligned}
\end{equation}
Here $\mathbf{T}_{m}$ and $\mathbf{T}_{v}$ are transition terms determined by the initial moment state at $t_{0}$, while $\mathbf{C}_{m}$ and $\mathbf{C}_{v}$ are curvature terms controlled by the variation of the logarithmic magnitude drift. More precisely, coordinate-wise,
\[
\begin{aligned}
|C_{m,i}(t)|
&\leq
B_{m,i}\tau_{1}^{2}(\Lambda^{2}+\Lambda'),\\
|C_{v,i}(t)|
&\leq
B_{v,i}\tau_{2}^{2}(\Lambda^{2}+\Lambda').
\end{aligned}
\]
for constants depending on the trajectory $\mathbf{g}$, the initial moments
$\mathbf{m}(t_{0})$ and $\mathbf{v}(t_{0})$, and the time scales
$\tau_{1}$ and $\tau_{2}$.
\end{proposition}

The first terms in \eqref{eq:moment-drift-decomposition} describe how the two exponential memories track the same coordinate-wise magnitude trajectory. Their explicit lag corrections have different coefficients:
$-\tau_{1}\bm{\delta}(t)$ in the first moment and
$-2\tau_{2}\bm{\delta}(t)$ in the second moment. The transition terms retain the effect of the state at $t_{0}$, while the curvature terms collect the dependence on $\bm{\delta}'$ and $\bm{\delta}^{2}$. Explicit expressions and the proof of Proposition~\ref{prop:m-v-delta} are given in Proposition~\ref{prop:m-v-delta-ap} of Appendix~\ref{appendix:technical-details}.

Passing from the moments to the normalized update reveals how the mismatch between the two memory times enters Adam. On a coordinate-wise sign-stable interval, the ratio
$\mathbf{m}(t)/\sqrt{\mathbf{v}(t)}$ separates into a sign component, an explicit magnitude-lag contribution, and the remaining transition, curvature, and nonlinear ratio effects.

\begin{theorem}
\label{thm:R-delta}
Let $I=[t_{0},t_{1}]$ be a compact interval and let
$\mathbf{g}:I\to\mathbb{R}^{d}$ be a $C^{2}$ mapping such that
$g_i(t)\neq 0$ for every $t\in I$ and every $i\in\{1,\ldots,d\}$.
Let $\bm{\delta}(t)$ be defined by \eqref{eq:delta-def}, and let
$\mathbf{m}(t)$ and $\mathbf{v}(t)$ solve the moment equations of the Adam flow. Then, at every time at which $ \mathbf{R}(t)
=
{\mathbf{m}(t)}/{\sqrt{\mathbf{v}(t)}} $ 
is defined,
\begin{equation}
\label{eq:R-drift-decomposition}
\mathbf{R}(t)
=
\operatorname{sign}\bigl(\mathbf{g}(t)\bigr)
\odot
\left(
\mathbf{1}
+
(\tau_{2}-\tau_{1})\bm{\delta}(t)
\right)
+
\mathbf{E}_{R}(t),
\end{equation}
where $\mathbf{E}_{R}(t)$ collects the transition, curvature, and nonlinear ratio terms induced by the decompositions of $\mathbf{m}(t)$ and $\mathbf{v}(t)$. A precise expression and bounds for $\mathbf{E}_{R}(t)$ are given in Theorem~\ref{thm:R-delta-ap} of Appendix~\ref{appendix:technical-details}.
\end{theorem}

\begin{remark}
\label{rem:R-remainder}
The decomposition in \eqref{eq:R-drift-decomposition} is exact. The remainder
$\mathbf{E}_{R}(t)$ retains the additional dependence carried by the transition, curvature, and nonlinear ratio effects. The term
\(
\operatorname{sign}\bigl(\mathbf{g}(t)\bigr)
\odot
(\tau_{2}-\tau_{1})\bm{\delta}(t)
\)
isolates the explicit magnitude-lag channel whose coefficient is determined solely by the mismatch between Adam's two memory times.
\end{remark}

\paragraph{Cancellation under tied memories.}
Define the explicit magnitude-lag contribution by
\[
\mathbf{L}(t)
:=
\operatorname{sign}\bigl(\mathbf{g}(t)\bigr)
\odot
(\tau_{2}-\tau_{1})\bm{\delta}(t).
\]
Its coefficient vanishes exactly when $\tau_{1}=\tau_{2}$. Hence, this channel is structurally absent for every admissible gradient trajectory if and only if the two memory times coincide. Under the parameterization
\(
\beta_{r}
=
e^{-\Delta t/\tau_{r}},
r\in\{1,2\},
\)
this condition is equivalent to $\beta_{1}=\beta_{2}$. In the tied-momentum regime, the normalized update therefore takes the form
\[
\mathbf{R}(t)
=
\operatorname{sign}\bigl(\mathbf{g}(t)\bigr)
+
\mathbf{E}_{R}(t),
\]
with $\mathbf{E}_{R}(t)$ retaining the history-dependent transition, curvature, and nonlinear ratio effects. Tying the memories thus changes the composition of Adam's update in one precise way: it removes the explicit response generated by a lag between the first- and second-moment filters.

\section{Empirical Signatures of the Magnitude-Lag Mechanism}
\label{sec:experiments}

The analysis in Section~\ref{sec:theory} identifies a concrete mechanism:
unequal memory times introduce the explicit magnitude-lag contribution
$\mathbf{L}$, while tying the memories removes it exactly. We test this
mechanism at three levels. A controlled gradient signal isolates the response
to changing magnitude. A literal full-history decomposition then measures the
theoretical terms on gradients from independently trained models. Finally,
full training runs examine whether the compositional change is accompanied by
a consistent difference in update-norm smoothness.

\subsection{Controlled Magnitude Variation}
\label{subsec:controlled-magnitude}

\begin{figure}[ht]
    \centering
    \begin{minipage}[t]{0.5\textwidth}
        \centering
        \includegraphics[width=\textwidth]{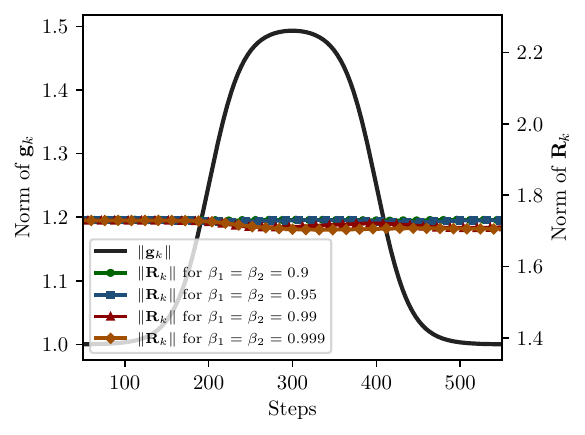}
        \captionof{figure}{Normalized-update response to a changing gradient magnitude for tied memories, $\beta_{1}=\beta_{2}$.}
        \label{fig:scale-experiment1}
    \end{minipage}
    \hfill
    \begin{minipage}[t]{0.5\textwidth}
        \centering
        \includegraphics[width=\textwidth]{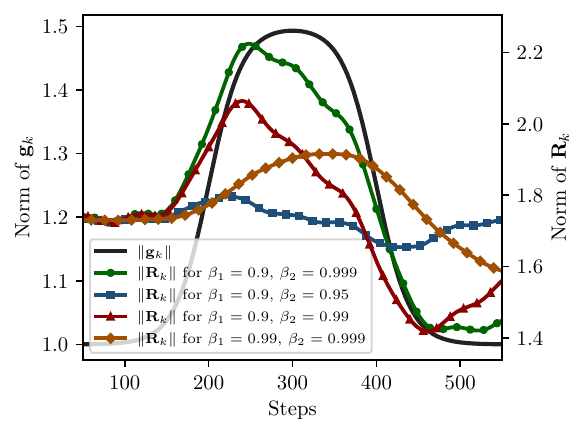}
        \captionof{figure}{Normalized-update response to the same gradient signal for untied memories, $\beta_{1}\neq\beta_{2}$.}
        \label{fig:scale-experiment2}
    \end{minipage}
\end{figure}

We begin with a sign-preserving gradient signal whose magnitude and direction vary over time. Replaying Adam on the same prescribed signal separates the effect of the memory parameters from changes in the optimization trajectory. Figures~\ref{fig:scale-experiment1} and~\ref{fig:scale-experiment2} show the gradient magnitude and the resulting normalized update for tied and untied memory configurations.

For tied memories, the update varies only through the remaining transition, curvature, nonlinear ratio, discretization, and $\varepsilon$ effects. Untied memories introduce the additional magnitude-lag channel identified in Theorem~\ref{thm:R-delta}, producing a visibly stronger response to the same variation in gradient magnitude. This controlled example illustrates the cancellation mechanism before evaluating its relevance on gradients obtained from training.

\subsection{A Literal Discrete Decomposition Along Training}
\label{subsec:decomposition-diagnostics}

We next test whether the terms in Theorem~\ref{thm:R-delta} are visible
along real optimization trajectories. We train a small MLP on Digits using
full-batch Adam and store the gradient before every update. Each
$(\beta_1,\beta_2)$ configuration is trained independently for three seeds;
the gradients are not synthetic and no counterfactual update is used. Full
protocol and robustness details are given in
Appendix~\ref{appendix:practical_validity}.

For a sign-stable coordinate, let
$u_{i,k}=\log|g_{i,k}|$ and let $\bar u_{\beta,i,k}$ denote its
bias-corrected exponential average. A first-order expansion of the discrete
Adam memories gives
\begin{equation}
 R_{i,k}
 =
 S_{i,k}+L^{\mathrm{disc}}_{i,k}+E_{i,k},
 \qquad
 S_{i,k}=\operatorname{sign}(g_{i,k}),
 \label{eq:discrete-R-decomposition}
\end{equation}
with
\begin{equation}
 \begin{aligned}
 L^{\mathrm{disc}}_{i,k}
 &:=S_{i,k}
 \left(\bar u_{\beta_1,i,k}-\bar u_{\beta_2,i,k}\right),\\
 E_{i,k}&:=R_{i,k}-S_{i,k}-L^{\mathrm{disc}}_{i,k}.
 \end{aligned}
 \label{eq:literal-discrete-lag}
\end{equation}
For a locally linear log-magnitude trajectory,
$\bar u_{\beta_1}-\bar u_{\beta_2}
\simeq(\ell_2-\ell_1)\delta$, where
$\ell_r=\beta_r/(1-\beta_r)$; hence
$L^{\mathrm{disc}}$ is the full-history discrete counterpart of the explicit
term in Theorem~\ref{thm:R-delta}. It vanishes identically on the diagonal and
contains no fitted parameter.

We retain coordinate-time samples that are nonzero and keep a constant sign
over the preceding $1000$ steps. The same window is used for every row and
the EMA states are never reset. We compute
$R=M_{\beta_1}/(\sqrt{V_{\beta_2}}+10^{-8})$ and report absolute component
budgets. We additionally report
\(
F_2=1-\sum E^2/\sum(R-S)^2
\), the fixed-prediction fraction of the off-diagonal deviation explained by
$L^{\mathrm{disc}}$.

\begin{table}[ht]
\centering
\caption{Literal discrete decomposition of Adam's normalized update on
independently trained real-gradient trajectories. Values are means over three
seeds.}
\label{tab:literal-discrete-decomposition}
\begin{tabular*}{\columnwidth}{@{\extracolsep{\fill}}lrrrr@{}}
\toprule
$(\beta_1,\beta_2)$ & $S$ & $L^{\rm disc}$ & $E$ & $F_2$ \\
\midrule
$(0.9,0.9)$     & $97.73\%$ & $0.00\%$  & $2.27\%$  & --      \\
$(0.9,0.99)$    & $62.52\%$ & $26.94\%$ & $10.54\%$ & $86.8\%$ \\
$(0.9,0.999)$   & $41.63\%$ & $45.01\%$ & $13.36\%$ & $64.3\%$ \\
$(0.99,0.99)$   & $74.37\%$ & $0.00\%$  & $25.63\%$ & --      \\
$(0.99,0.999)$  & $46.68\%$ & $43.12\%$ & $10.20\%$ & $83.8\%$ \\
$(0.999,0.999)$ & $75.42\%$ & $0.00\%$  & $24.58\%$ & --      \\
\bottomrule
\end{tabular*}
\end{table}

Table~\ref{tab:literal-discrete-decomposition} displays the structural change
predicted by the theory. With $\beta_1=0.9$, moving from the diagonal to
$\beta_2=0.99$ and $0.999$ reduces the sign budget from $97.73\%$ to
$62.52\%$ and $41.63\%$, while the lag budget grows from zero to $26.94\%$
and $45.01\%$. The residual remains between $10.20\%$ and $13.36\%$ in all
off-diagonal rows. Moreover, $L^{\mathrm{disc}}$ has the same sign as $R-S$ in more than $99\%$
of valid evaluations. Using the much shorter window $w=6$ still gives
off-diagonal residual budgets of only $14$--$16\%$
(Appendix~\ref{appendix:practical_validity}). Thus the observed decomposition
is not created by the long-window selection: tying the memories removes a
substantial full-history magnitude-lag term and shifts the realized update
toward its sign component.

\subsection{Update-Norm Dynamics Across Training Tasks}
\label{subsec:update-norm-dynamics}

We finally examine the aggregate training footprint of this compositional change. A sign component has fixed magnitude in every active coordinate, whereas the magnitude-lag and remainder terms introduce coordinate-wise amplitude variation. This motivates testing whether tied configurations, which remove
$\mathbf{L}$ and yield the sign-dominated decompositions in
Table~\ref{tab:literal-discrete-decomposition}, are accompanied by smoother
trajectories of the normalized-update norm.

Let
\(
q_k
:=
\operatorname{EMA}_{200}\!\left(\|\mathbf{R}_k\|\right)
\)
denote the update norm after exponential smoothing with a window of $200$ steps. We summarize its step-to-step variation by
\[
\omega
:=
\frac{1}{K-1}
\sum_{k=1}^{K-1}
|q_{k+1}-q_k|.
\]
The role of $\omega$ is to measure observable update-norm smoothness. The
literal decomposition in Section~\ref{subsec:decomposition-diagnostics}
directly measures the internal mismatch mechanism. Robustness of that
decomposition and of the oscillation analysis is reported in
Appendices~\ref{appendix:practical_validity} and
\ref{appendix:additional_experiments}, respectively.

Experiments are conducted on six model--dataset pairs covering both
vision and language tasks:
NanoGPT~\citep{karpathy2022nanogpt} on SlimPajama~\citep{soboleva2023slimpajama} 
and WikiText~\citep{merity2016pointer},
EfficientNet-B0~\citep{tan2019efficientnet} on TinyImageNet~\citep{le2015tinyimagenet},
ResNet18~\citep{he2016deep} and ViT-B16~\citep{dosovitskiy2021image} on
CIFAR-100~\citep{krizhevsky2009learning},
and T5~\citep{raffel2020exploring} on SQuAD~\citep{rajpurkar2016squad}.
For each pair, we train the model using Adam or AdamW over a $3\times3$
grid of momentum parameters
$\beta_{1},\beta_{2}\in\{0.9,0.99,0.999\}$,
except for ResNet18 on CIFAR-100, where we use a finer $5\times5$ grid
$\beta_{1},\beta_{2}\in\{0.9,0.968,0.99,0.9968,0.999\}$
to obtain a more detailed view of the same effect.
All configurations are repeated with three different random seeds, except
for NanoGPT on SlimPajama, where a single seed is used due to computational
constraints; see Appendix~\ref{appendix:implementation_details} for full
implementation and training details.

Table \ref{tab:omega_grid_real} summarizes the oscillation values $\omega=\omega(\beta_{1},\beta_{2})$
across all experiments. For each fixed $\beta_{1}$, the minimum oscillation
is typically attained on the diagonal $\beta_{2}=\beta_{1}$. This trend is further quantified by the
\emph{Rate} column, which reports the fraction of cases, fixing
$\beta_{1}$ and the random seed, in which the minimizing $\beta_{2}$
coincides with $\beta_{1}$.

\begin{table*}[ht]
\centering
\setlength{\tabcolsep}{3.0pt}
\renewcommand{\arraystretch}{1.02}
\caption{%
Oscillation values \(\omega(\beta_{1},\beta_{2})\), averaged over seeds when applicable.
Bold entries mark the minimum within each fixed-\(\beta_1\) row.
\emph{Rate} is the fraction of trials where the minimizing \(\beta_2\) equals \(\beta_1\).}
\label{tab:omega_grid_real}
\begin{tabular*}{\textwidth}{
  @{\extracolsep{\fill}}
  c c c c c c c c
  @{}
}
\toprule
\multirow[c]{2}{*}{Model/Dataset} &
\multirow[c]{2}{*}{$\beta_1$} &
\multicolumn{5}{c}{$\omega(\beta_1,\beta_2)$} &
\multirow[c]{2}{*}{Rate} \\
\cmidrule(lr){3-7}
& &
\multicolumn{2}{c}{$\beta_2=0.9$} &
$\beta_2=0.99$ &
\multicolumn{2}{c}{$\beta_2=0.999$} &
\\
\midrule

\multirow[c]{3}{*}{\begin{tabular}{c}NanoGPT\\WikiText\end{tabular}} &
$0.9$   & \multicolumn{2}{c}{\textbf{0.6329}} & 1.063  & \multicolumn{2}{c}{1.147} &
\multirow[c]{3}{*}{100\%} \\
& $0.99$  & \multicolumn{2}{c}{0.4227} & \textbf{0.2413} & \multicolumn{2}{c}{0.2644} & \\
& $0.999$ & \multicolumn{2}{c}{0.5249} & 0.2356 & \multicolumn{2}{c}{\textbf{0.05768}} & \\
\midrule

\multirow[c]{3}{*}{\begin{tabular}{c}EfficientNet-B0\\TinyImageNet\end{tabular}} &
$0.9$   & \multicolumn{2}{c}{\textbf{0.08169}} & 0.1869 & \multicolumn{2}{c}{0.217} &
\multirow[c]{3}{*}{100\%} \\
& $0.99$  & \multicolumn{2}{c}{$3.312\times 10^{5}$} & \textbf{0.01430} & \multicolumn{2}{c}{0.03121} & \\
& $0.999$ & \multicolumn{2}{c}{$4.918\times 10^{6}$} & $2.466\times 10^{5}$ & \multicolumn{2}{c}{\textbf{0.002931}} & \\
\midrule

\multirow[c]{3}{*}{\begin{tabular}{c}T5\\SQuAD\end{tabular}} &
$0.9$   & \multicolumn{2}{c}{0.3902} & \textbf{0.3455} & \multicolumn{2}{c}{0.5714} &
\multirow[c]{3}{*}{77.8\%} \\
& $0.99$  & \multicolumn{2}{c}{1.203} & \textbf{0.03733} & \multicolumn{2}{c}{0.06729} & \\
& $0.999$ & \multicolumn{2}{c}{1.731} & 1.137 & \multicolumn{2}{c}{\textbf{0.003777}} & \\
\midrule

\multirow[c]{3}{*}{\begin{tabular}{c}NanoGPT\\SlimPajama\end{tabular}} &
$0.9$   & \multicolumn{2}{c}{\textbf{0.4988}} & 0.8467 & \multicolumn{2}{c}{1.169} &
\multirow[c]{3}{*}{100\%} \\
& $0.99$  & \multicolumn{2}{c}{0.3404} & \textbf{0.1947} & \multicolumn{2}{c}{0.2489} & \\
& $0.999$ & \multicolumn{2}{c}{0.3372} & 0.07359 & \multicolumn{2}{c}{\textbf{0.05921}} & \\
\midrule

\multirow[c]{3}{*}{\begin{tabular}{c}ViT-B16\\CIFAR-100\end{tabular}} &
$0.9$   & \multicolumn{2}{c}{\textbf{0.2925}} & 0.641 & \multicolumn{2}{c}{0.6777} &
\multirow[c]{3}{*}{100\%} \\
& $0.99$  & \multicolumn{2}{c}{NaN} & \textbf{0.07104} & \multicolumn{2}{c}{0.09244} & \\
& $0.999$ & \multicolumn{2}{c}{NaN} & 204.2 & \multicolumn{2}{c}{\textbf{0.0735}} & \\

\specialrule{0.8pt}{2pt}{2pt}
& &
$\beta_2=0.9$ &
$\beta_2=0.968$ &
$\beta_2=0.99$ &
$\beta_2=0.9968$ &
$\beta_2=0.999$ &
\\
\midrule

\multirow[c]{5}{*}{\begin{tabular}{c}ResNet18\\CIFAR-100\end{tabular}} &
$0.9$  & 0.1263 & \textbf{0.0915} & 0.1032 & 0.1443 & 0.1512 &
\multirow[c]{5}{*}{80\%} \\
& $0.968$  & 0.0300 & \textbf{0.0264} & 0.0433 & 0.0487 & 0.0566 & \\
& $0.99$  & 0.6238 & 0.0126 & \textbf{0.0087} & 0.0137 & 0.0154 & \\
& $0.9968$ & 1.6459 & 0.3734 & 0.0073 & \textbf{0.0038} & 0.0046 & \\
& $0.999$  & 0.8606 & 0.5543 & 0.1650 & 0.0050 & \textbf{0.0022} & \\

\bottomrule
\end{tabular*}
\end{table*}
This pattern is consistent with the literal decomposition in
Table~\ref{tab:literal-discrete-decomposition}. On the diagonal, the explicit
magnitude-lag channel vanishes and the measured update is sign-dominated.
Training-dynamics plots for all six tasks are reported in
Appendix~\ref{appendix:additional_experiments}, showing the same qualitative
pattern.

\begin{remark}
Update-norm smoothness and validation performance quantify different aspects of an optimization trajectory. Validation performance additionally depends on learning-rate calibration, stochastic noise, regularization, batch size, and the interaction between the optimizer and the task. Consequently, the configuration with the smallest $\omega$ can differ from the configuration with the best validation metric on a finite grid. The experiments here characterize the update dynamics associated with tied memories; broader comparisons of predictive performance across diagonal settings are reported by \citet{AdamSecretSauce}.
\end{remark}

Together, the three diagnostics in Section~\ref{sec:experiments} connect
the local decomposition to full training behavior. Controlled magnitude
variation exposes the additional response introduced by unequal memory
times. The literal discrete decomposition recovers the predicted transfer of
weight from the sign term to the magnitude-lag term away from the diagonal,
with a comparatively small residual. Across full training runs, tied
configurations typically exhibit smoother update-norm trajectories. These
observations provide a consistent empirical signature of the
cancellation identified in Section~\ref{sec:theory}.

\section{Conclusions}

We have identified a concrete mechanism that makes tied momentum dynamically special in Adam. On coordinate-wise sign-stable intervals, memory mismatch creates an explicit magnitude-lag contribution proportional to $(\tau_{2}-\tau_{1})\,d\log|g|/dt$, alongside transition, curvature, and nonlinear ratio effects. This channel is structurally absent exactly on the diagonal $\beta_{1}=\beta_{2}$, while the remaining history-dependent dynamics are retained.

A literal full-history decomposition on real training gradients shows that
the sign term dominates on the diagonal, while the predicted magnitude-lag
term becomes substantial off it and leaves a comparatively small residual.
Across six vision and language tasks, tied configurations typically exhibit
the corresponding smoother-update signature. Thus, memory-scale mismatch is
a concrete source of magnitude sensitivity in Adam, and tying the memories
uniquely removes that channel.

\section*{AI Use Statement}

Generative AI tools were used as interactive research aids during the development of parts of the mathematical analysis, including exploring proof strategies, deriving and checking intermediate expressions, and discussing formulations and interpretations of theoretical claims. They were also used for language editing and presentation feedback. All AI-assisted mathematical material was critically examined by the authors, re-derived or corrected where necessary, and integrated only after careful verification. The authors verified the final mathematical statements, proofs, experiments, and interpretations and take full responsibility for the content of the paper.

\flushcolsend
\clearpage

\bibliography{bibliography}
\bibliographystyle{plainnat}

\flushcolsend
\clearpage

\onecolumn
\appendix
\section*{Future Work}

A first direction is to understand what happens beyond the sign-stable
regime analyzed in this work. Sign changes are not merely a technical
nuisance: they mark coordinates where the optimizer reverses the
direction of motion. From the viewpoint of the present theory, these are
also the points where the logarithmic magnitude drift becomes singular, since
$\log |g(t)|$ is not well behaved through zero crossings. Extending the
analysis across such events may require a representation that tracks
magnitude together with sign, or more generally magnitude together with
phase. This would complement the present sign-stable analysis, which
describes the intervals where the gradient keeps a fixed sign.

A second direction is to refine the structure of the remaining term
$\mathbf{E}_R$. The current decomposition isolates the explicit
magnitude-lag channel while retaining transition, curvature, and nonlinear
ratio effects inside $\mathbf{E}_R$. Identifying regimes in which these
contributions can be separated or controlled more sharply could clarify when
Adam becomes strongly sign-dominated and provide more precise principles for
optimizer design.

An additional limitation concerns the interpretation of $\omega$. In the
literal discrete decomposition, tied-memory updates are sign-dominated on
long sign-stable intervals, which motivates examining the smoothness of the
aggregate update norm.
However, $\omega$ does not identify the internal decomposition, and a small
value may arise through other mechanisms. Developing diagnostics that connect
the component-wise and aggregate dynamics more directly is therefore a natural
direction for future work.

More broadly, the perspective developed in this work suggests that
the structured processing of gradient-magnitude information provides a useful
principle for understanding and designing modern optimizers.

\section{Technical Details for the Adam Flow}
\label{appendix:technical-details}

This appendix collects the technical material that supports the statements in the main text. It opens with a short derivation of the continuous-time Adam flow from the discrete updates, clarifying the role of the relaxation times $\tau_1$ and $\tau_2$. It then turns to a detailed analysis of the behavior of the first and second moments $\mathbf{m}(t)$ and $\mathbf{v}(t)$, and of their relation to the logarithmic magnitude drift $\bm{\delta}(t)$.

\subsection{From Discrete Adam to the Continuous-time Flow}\label{appendix:continuum}
This first section justifies the continuous-time system \eqref{eq:adam-flow} and clarifies the role of the parameters $\beta_1,\beta_2$ and $\tau_1,\tau_2$.

Consider, for simplicity, a single coordinate of Adam’s first moment. The discrete update is
\[
m_{k+1} = \beta_1 m_k + (1-\beta_1)\, g_k,
\qquad t_k = k\,\Delta t .
\]
Assuming that $m_k \approx m(t_k)$ and $g_k \approx g(t_k)$ for smooth functions $m(t)$ and $g(t)$, the forward difference can be written as
\[
\frac{m_{k+1}-m_k}{\Delta t}
= \frac{\beta_1-1}{\Delta t}\, m_k + \frac{1-\beta_1}{\Delta t}\, g_k .
\]
When $\beta_1$ is close to $1$, as is the case in practical implementations of Adam, a relaxation time $\tau_1>0$ may be introduced through the asymptotic relation
\[
1-\beta_1 = \frac{\Delta t}{\tau_1} + o(\Delta t).
\]
This implies
\[
\frac{\beta_1-1}{\Delta t} = -\frac{1}{\tau_1} + o(1),
\qquad
\frac{1-\beta_1}{\Delta t} = \frac{1}{\tau_1} + o(1).
\]
Letting $\Delta t\to 0$ while keeping $\tau_1$ fixed, the difference quotient converges to $m'(t)$, yielding
\[
\tau_1 m'(t) = - m(t) + g(t),
\]
which recovers the first equation in \eqref{eq:adam-flow}. The same argument applied to
\[
v_{k+1} = \beta_2 v_k + (1-\beta_2)\, g_k^2
\]
gives
\[
\tau_2 v'(t) = - v(t) + g(t)^2.
\]

An important point is that the correspondence between $\beta$ and $\tau$ is not unique. Any relation of the form
\[
\beta(\tau, \Delta t) = 1 - \frac{\Delta t}{\tau} + o(\Delta t)
\]
leads to the same limiting differential equation. The exponential parametrization
\[
\beta = e^{-\Delta t/\tau}
\]
adopted in the main text is a convenient choice, as it enforces $0<\beta<1$ for all $\Delta t>0$ and makes the small--$\Delta t$ behavior explicit, but it is by no means the only one compatible with the continuous-time limit.

This perspective also clarifies the interpretation of the parameters. Although one may write $\beta=\beta(\tau,\Delta t)$ through relations such as $\beta=e^{-\Delta t/\tau}$, in Adam the quantities $\beta_1$ and $\beta_2$ are the primitive, fixed hyperparameters of the algorithm. The corresponding time scales $\tau_1$ and $\tau_2$ are therefore defined implicitly from $(\beta,\Delta t)$ and represent the effective relaxation times of the associated continuous-time system. In this sense, $\tau_1$ and $\tau_2$ characterize the limiting continuous dynamics, whereas $\beta_1$ and $\beta_2$ are the discrete parameters appearing in Adam’s updates.

\subsection{Tracking Decompositions for the Moments
\texorpdfstring{$\mathbf{m}(t)$ and $\mathbf{v}(t)$}{m(t) and v(t)}}
\label{appendix:mv-decomposition}

Recall that the Adam flow is given by
\begin{equation}
\label{eq:adam-flow-app}
\begin{aligned}
\tau_{1} \mathbf{m}'(t) &= - \mathbf{m}(t) + \mathbf{g}(t),\\
\tau_{2} \mathbf{v}'(t) &= - \mathbf{v}(t) + \mathbf{g}(t)^{ 2},\\
\boldsymbol{\theta}'(t) &= - \bar{\eta}\,\frac{\mathbf{m}(t)}{\sqrt{\mathbf{v}(t)}},
\end{aligned}
\end{equation}
where the operations are understood coordinate-wise, and that the
logarithmic magnitude drift is defined by
\begin{equation}
\label{eq:delta-def-app}
\delta_{i}(t) := \frac{d}{dt}\log|g_{i}(t)|
= \frac{ g'_{i}(t)}{g_{i}(t)},
\end{equation}
for each coordinate $i$, whenever $g_{i}(t)\neq 0$.
We write $\bm{\delta}(t) = (\delta_{1}(t),\dots,\delta_{d}(t))$.

To streamline the proof of Proposition~\ref{prop:m-v-delta}, it is
convenient to isolate an exact decomposition for the scalar linear ODE
$\tau x'=-x+y(t)$. The decomposition separates the filtered state into
an instantaneous tracking term, an explicit lag term, a transition term
depending on the initial state, and a curvature term depending on
$y''$.

\begin{lemma}
\label{lem:tracking}
Let $I=[t_{0},t_{1}]$ be a compact interval and let
$y \in C^{2}(I,\mathbb{R})$.
For fixed $\tau>0$, let $x:I\to\mathbb{R}$ solve
\begin{equation*}
\tau x'(t) = - x(t) + y(t),
\end{equation*}
for $t\in I$, with initial condition $x(t_{0})=x_{0}$. Then, for all $t\in I$,
\[
x(t)=y(t)-\tau y'(t)+T_x(t)+C_x(t),
\]
where
\[
T_x(t)
=
\bigl(x_0-y(t_0)+\tau y'(t_0)\bigr)e^{-(t-t_0)/\tau}
\]
and
\[
C_x(t)
=
\tau^2\int_{t_0}^{t}K(t,s)y''(s)\,ds.
\]
Moreover, if $M_y:=\sup_{t\in I}|y''(t)|$, then
\[
|C_x(t)|\leq \tau^2 M_y.
\]
\end{lemma}

\begin{proof}
The solution of $\tau x'(t) = -x(t) + y(t)$ with initial condition $x(t_0)= x_{0}$ follows from the variation of constants formula and reads
\[
x(t) = e^{-(t-t_{0})/\tau} x_{0}
+ \int_{t_{0}}^{t} K(t,s)\,y(s)\,ds
\]
with 
\[
K(t,s) := \frac{1}{\tau} e^{-(t-s)/\tau},
\]
Notice that
\[
\partial_{s} K(t,s) = \frac{1}{\tau^{2}} e^{-(t-s)/\tau}
= \frac{1}{\tau} K(t,s),
\]
so that an integration by parts yields
\[
\begin{aligned}
\int_{t_{0}}^{t} K(t,s)\,y(s)\,ds
&=  \tau K(t,t) y(t) - \tau K(t,t_{0}) y(t_{0})
- \tau \int_{t_{0}}^{t} K(t,s)\, y'(s)\,ds\\
& =  y(t) - e^{-(t-t_{0})/\tau} y(t_{0})
- \tau \int_{t_{0}}^{t} K(t,s)\, y'(s)\,ds.
\end{aligned}
\]
Substituting the expression of $x(t)$ implies that 
\[
\begin{aligned}
x(t)
&= e^{-(t-t_{0})/\tau} x_{0}
+ y(t) - e^{-(t-t_{0})/\tau} y(t_{0})
- \tau \int_{t_{0}}^{t} K(t,s)\, y'(s)\,ds.
\end{aligned}
\]

Applying integration by parts once again to the term involving $y'$ yields
\[
\begin{aligned}
\tau \int_{t_{0}}^{t} K(t,s)\, y'(s)\,ds
&=  \tau^2  K(t,t)y'(t) - \tau^2 K(t,t_{0})y'(t_{0})
- \tau^{2} \int_{t_{0}}^{t} K(t,s)\,y''(s)\,ds\\
&=  \tau y'(t) - \tau e^{-(t-t_{0})/\tau}y'(t_{0})
- \tau^{2} \int_{t_{0}}^{t} K(t,s)\,y''(s)\,ds.
\end{aligned}
\]
Combining the expressions obtained above, it follows that
\[
\begin{aligned}
x(t)
&= e^{-(t-t_{0})/\tau} x_{0}
+ y(t) - e^{-(t-t_{0})/\tau} y(t_{0})
- \tau y'(t) + \tau e^{-(t-t_{0})/\tau}y'(t_{0}) + \tau^{2} \int_{t_{0}}^{t} K(t,s)\,y''(s)\,ds\\
&= y(t) - \tau y'(t)
+ \Bigl(x_{0} - y(t_{0}) + \tau y'(t_{0})\Bigr)e^{-(t-t_{0})/\tau}
+ \tau^{2} \int_{t_{0}}^{t} K(t,s)\,y''(s)\,ds,
\end{aligned}
\]
which yields the desired representation with
\[
T_x(t) =
\Bigl(x_{0} - y(t_{0}) + \tau y'(t_{0})\Bigr)e^{-(t-t_{0})/\tau} \quad \text{and} \quad C_x(t) =  \tau^{2} \int_{t_{0}}^{t} K(t,s)\,y''(s)\,ds.
\]
Finally, as $K(t,s)\ge 0$ and
\[
\int_{t_{0}}^{t} K(t,s)\,ds
= \int_{0}^{t-t_{0}} \frac{1}{\tau} e^{-u/\tau}\,du \le 1,
\]
it follows that
\[
\left|C_x(t)\right|
\le \tau^{2} M_y,
\]
which produces the bound of the statement.
\end{proof}

With this lemma in place, the proposition describing the tracking behavior of Adam’s moments can be established.

\begin{proposition}[Tracking decomposition of $\mathbf{m}$ and $\mathbf{v}$]
\label{prop:m-v-delta-ap}
Let $I=[t_{0},t_{1}]$ be a compact interval and let 
$\mathbf{g}:I\to\mathbb{R}^{d}$ be a $C^{2}$ mapping such that, for each coordinate 
$i\in\{1,\dots,d\}$, one has $g_i(t)\neq 0$ for all $t\in I$. Let
\[
\Lambda := \sup_{t\in I}\max_{1\le i\le d}|\delta_i(t)|,
\qquad
\Lambda' := \sup_{t\in I}\max_{1\le i\le d}|\delta'_i(t)|.
\]
Let $\mathbf{m}(t)$ and $\mathbf{v}(t)$ solve the first two equations of the Adam flow
\eqref{eq:adam-flow-app} on $I$, with arbitrary initial conditions at $t_0$.
Then, for each coordinate $i$ and all $t\in I$,
\[
m_i(t)
=
g_i(t)\bigl(1-\tau_1\delta_i(t)\bigr)
+
T_{m,i}(t)
+
C_{m,i}(t),
\]
and
\[
v_i(t)
=
g_i(t)^2\bigl(1-2\tau_2\delta_i(t)\bigr)
+
T_{v,i}(t)
+
C_{v,i}(t),
\]
where the transition terms are
\[
T_{m,i}(t)
=
\Bigl(
m_i(t_0)-g_i(t_0)+\tau_1 g_i(t_0)\delta_i(t_0)
\Bigr)
e^{-(t-t_0)/\tau_1},
\]
and
\[
T_{v,i}(t)
=
\Bigl(
v_i(t_0)-g_i(t_0)^2
+
2\tau_2 g_i(t_0)^2\delta_i(t_0)
\Bigr)
e^{-(t-t_0)/\tau_2},
\]
while the curvature terms are
\[
C_{m,i}(t)
=
\tau_1^2
\int_{t_0}^{t}
K_1(t,s)\,
g_i(s)
\bigl(\delta_i(s)^2+\delta_i'(s)\bigr)
\,ds,
\]
and
\[
C_{v,i}(t)
=
\tau_2^2
\int_{t_0}^{t}
K_2(t,s)\,
2g_i(s)^2
\bigl(2\delta_i(s)^2+\delta_i'(s)\bigr)
\,ds,
\]
with
\[
K_j(t,s)=\frac{1}{\tau_j}e^{-(t-s)/\tau_j},
\qquad j\in\{1,2\}.
\]
Moreover, the transition terms satisfy
\[
|T_{m,i}(t)|
\le
A_{m,i}e^{-(t-t_0)/\tau_1},
\qquad
|T_{v,i}(t)|
\le
A_{v,i}e^{-(t-t_0)/\tau_2},
\]
where
\[
A_{m,i}
=
\Bigl|
m_i(t_0)-g_i(t_0)+\tau_1 g_i(t_0)\delta_i(t_0)
\Bigr|,
\]
and
\[
A_{v,i}
=
\Bigl|
v_i(t_0)-g_i(t_0)^2
+
2\tau_2 g_i(t_0)^2\delta_i(t_0)
\Bigr|.
\]
The curvature terms satisfy
\[
|C_{m,i}(t)|
\le
B_{m,i}\tau_1^2(\Lambda^2+\Lambda'),
\qquad
|C_{v,i}(t)|
\le
B_{v,i}\tau_2^2(\Lambda^2+\Lambda'),
\]
with
\[
B_{m,i}=\sup_{t\in I}|g_i(t)|,
\qquad
B_{v,i}=4\sup_{t\in I}|g_i(t)|^2.
\]
\end{proposition}
\begin{proof}
Since the dynamics decouples across coordinates, fix an index $i$ and
work on a single component, denoted by $g(t)$, $m(t)$, $v(t)$ and
$\delta(t)$ to simplify notation.

\medskip\noindent
\emph{Step 1: decomposition for $m(t)$.}
The first equation of the Adam flow reads
\[
\tau_{1} m'(t) = - m(t) + g(t).
\]
Applying Lemma~\ref{lem:tracking} with $y(t)=g(t)$ and $\tau=\tau_{1}$
yields
\[
m(t)
=
g(t)-\tau_{1}g'(t)
+
T_m(t)
+
C_m(t),
\]
where
\[
T_m(t)
=
\Bigl(
m(t_0)-g(t_0)+\tau_1 g'(t_0)
\Bigr)
e^{-(t-t_0)/\tau_1}
\]
and
\[
C_m(t)
=
\tau_1^2
\int_{t_0}^{t}
K_1(t,s)g''(s)\,ds,
\qquad
K_1(t,s)
=
\frac{1}{\tau_1}e^{-(t-s)/\tau_1}.
\]
By definition, $\delta(t)=g'(t)/g(t)$, hence
\[
g'(t)=g(t)\delta(t)
\quad\text{and}\quad
g''(t)=\bigl(\delta(t)^2+\delta'(t)\bigr)g(t).
\]
Substituting $g'(t)=g(t)\delta(t)$ into the expression for $m(t)$ gives
\[
m(t)
=
g(t)\bigl(1-\tau_1\delta(t)\bigr)
+
T_m(t)
+
C_m(t),
\]
with
\[
T_m(t)
=
\Bigl(
m(t_0)-g(t_0)+\tau_1 g(t_0)\delta(t_0)
\Bigr)
e^{-(t-t_0)/\tau_1}
\]
and
\[
C_m(t)
=
\tau_1^2
\int_{t_0}^{t}
K_1(t,s)
g(s)
\bigl(\delta(s)^2+\delta'(s)\bigr)
\,ds.
\]
Moreover, since $K_1(t,s)\ge 0$ and
\[
\int_{t_0}^{t}K_1(t,s)\,ds\le 1,
\]
if $B:=\sup_{t\in I}|g(t)|$, then
\[
|C_m(t)|
\le
\tau_1^2
B
(\Lambda^2+\Lambda').
\]
The transition term satisfies
\[
|T_m(t)|
\le
A_m e^{-(t-t_0)/\tau_1},
\]
where
\[
A_m
=
\Bigl|
m(t_0)-g(t_0)+\tau_1 g(t_0)\delta(t_0)
\Bigr|.
\]
This proves the stated decomposition and bounds for $m_i(t)$.

\medskip\noindent
\emph{Step 2: decomposition for $v(t)$.}
The second equation of the Adam flow is
\[
\tau_{2} v'(t) = - v(t) + g(t)^2.
\]
Lemma~\ref{lem:tracking} applies again, now with $y(t)=g(t)^2$ and
$\tau=\tau_2$, yielding
\[
v(t)
=
g(t)^2-\tau_2(g^2)'(t)
+
T_v(t)
+
C_v(t),
\]
where
\[
T_v(t)
=
\Bigl(
v(t_0)-g(t_0)^2+\tau_2(g^2)'(t_0)
\Bigr)
e^{-(t-t_0)/\tau_2}
\]
and
\[
C_v(t)
=
\tau_2^2
\int_{t_0}^{t}
K_2(t,s)(g^2)''(s)\,ds,
\qquad
K_2(t,s)
=
\frac{1}{\tau_2}e^{-(t-s)/\tau_2}.
\]
Since
\[
(g^2)'(t)=2g(t)g'(t)=2g(t)^2\delta(t),
\]
and
\[
\begin{aligned}
(g^2)''(t)
&=
2\bigl(g'(t)\bigr)^2
+
2g(t)g''(t)\\
&=
2g(t)^2\delta(t)^2
+
2g(t)^2\bigl(\delta'(t)+\delta(t)^2\bigr)\\
&=
2g(t)^2\bigl(2\delta(t)^2+\delta'(t)\bigr),
\end{aligned}
\]
we obtain
\[
v(t)
=
g(t)^2
\bigl(1-2\tau_2\delta(t)\bigr)
+
T_v(t)
+
C_v(t),
\]
with
\[
T_v(t)
=
\Bigl(
v(t_0)-g(t_0)^2
+
2\tau_2g(t_0)^2\delta(t_0)
\Bigr)
e^{-(t-t_0)/\tau_2}
\]
and
\[
C_v(t)
=
\tau_2^2
\int_{t_0}^{t}
K_2(t,s)
2g(s)^2
\bigl(2\delta(s)^2+\delta'(s)\bigr)
\,ds.
\]
Again, since $K_2(t,s)\ge 0$ and
\[
\int_{t_0}^{t}K_2(t,s)\,ds\le 1,
\]
if $B:=\sup_{t\in I}|g(t)|$, then
\[
|C_v(t)|
\le
4\tau_2^2B^2(\Lambda^2+\Lambda').
\]
The transition term satisfies
\[
|T_v(t)|
\le
A_v e^{-(t-t_0)/\tau_2},
\]
where
\[
A_v
=
\Bigl|
v(t_0)-g(t_0)^2
+
2\tau_2g(t_0)^2\delta(t_0)
\Bigr|.
\]
This proves the stated decomposition and bounds for $v_i(t)$, and the result follows.
\end{proof}
The proposition above gives the explicit version of the decomposition
used in the main text. The Adam moments are separated into an
instantaneous tracking term, a magnitude-lag term proportional to the
logarithmic drift, a transition term determined by the initial moment
state, and a curvature term controlled by the variation of the drift. In
the empirical diagnostics, these contributions are measured separately
rather than absorbed into a single asymptotic remainder.

\subsection{Decomposition of the Normalized Update
\texorpdfstring{$\mathbf{R}(t)$}{R(t)}}
\label{appendix:R-decomposition}

Building on the decompositions of $\mathbf{m}(t)$ and $\mathbf{v}(t)$ obtained in
Proposition~\ref{prop:m-v-delta-ap}, we now derive a corresponding
decomposition for the normalized update
\[
\mathbf{R}(t) := \frac{\mathbf{m}(t)}{\sqrt{\mathbf{v}(t)}}.
\]
The purpose is to isolate the explicit magnitude-lag term in the Adam update,
while keeping the transition, curvature, and nonlinear ratio effects explicit.

\begin{theorem}[Decomposition of the normalized update]
\label{thm:R-delta-ap}
Let $I=[t_{0},t_{1}]$ be a compact interval and let
$\mathbf{g}:I\to\mathbb{R}^{d}$ be a $C^{2}$ mapping such that, for each
coordinate $i\in \{1,\dots,d\}$, one has $g_i(t)\neq 0$ for all $t\in I$.
Let $\mathbf{m}(t)$ and $\mathbf{v}(t)$ solve the first two equations
of the Adam flow \eqref{eq:adam-flow-app} on $I$, with Adam-valid second
moment initialization $v_i(t_0)\ge 0$.

Let $T_{m,i},C_{m,i},T_{v,i},C_{v,i}$ be the transition and curvature terms
from Proposition~\ref{prop:m-v-delta-ap}. Then, for each coordinate $i$
and every $t\in(t_0,t_1]$,
\[
R_i(t)
=
\operatorname{sign}\bigl(g_i(t)\bigr)
\Bigl(
1+(\tau_2-\tau_1)\delta_i(t)
\Bigr)
+
E_{R,i}(t),
\]
where $E_{R,i}(t)$ contains the transition, curvature, and nonlinear ratio
terms. More precisely, if
\[
q_{m,i}(t)
:=
\frac{T_{m,i}(t)+C_{m,i}(t)}{g_i(t)},
\qquad
q_{v,i}(t)
:=
\frac{T_{v,i}(t)+C_{v,i}(t)}{g_i(t)^2},
\]
then
\[
E_{R,i}(t)
=
\operatorname{sign}\bigl(g_i(t)\bigr)
\left(
q_{m,i}(t)
-
\frac12 q_{v,i}(t)
+
N_{R,i}(t)
\right),
\]
where the nonlinear ratio term satisfies, on every compact subinterval
$J\subset(t_0,t_1]$,
\[
|N_{R,i}(t)|
\le
C_{R,i,J}
\left(
\tau_*|\delta_i(t)|
+
|q_{m,i}(t)|
+
|q_{v,i}(t)|
\right)^2,
\qquad t\in J,
\]
with $\tau_*=\max\{\tau_1,\tau_2\}$. Consequently, on such $J$,
\[
|E_{R,i}(t)|
\le
|q_{m,i}(t)|
+
\frac12 |q_{v,i}(t)|
+
C_{R,i,J}
\left(
\tau_*|\delta_i(t)|
+
|q_{m,i}(t)|
+
|q_{v,i}(t)|
\right)^2.
\]
\end{theorem}

\begin{proof}
Fix a coordinate index $i$ and omit the subscript $i$ from the notation:
the scalar quantities associated with this coordinate are denoted by
$g(t)$, $m(t)$, $v(t)$, $\delta(t)$, and $R(t)$.

Since $g$ is continuous and nonzero on the compact interval $I$, its sign
is constant on $I$ and there exists $g_{\min}>0$ such that
$|g(t)|\ge g_{\min}$ for all $t\in I$.

\medskip\noindent
\emph{Step 1: positivity of $v(t)$.}
The variation-of-constants formula for $\tau_{2}v'(t)=-v(t)+g(t)^2$
gives, for all $t\in I$,
\[
v(t)
=
e^{-(t-t_0)/\tau_2}v(t_0)
+
\frac{1}{\tau_2}
\int_{t_0}^{t}
e^{-(t-s)/\tau_2}g(s)^2\,ds .
\]
Since $v(t_0)\ge 0$ and $g(s)^2\ge g_{\min}^2$ on $I$, we obtain
\[
v(t)
\ge
g_{\min}^2
\left(
1-e^{-(t-t_0)/\tau_2}
\right),
\]
for every $t\in[t_0,t_1]$. Hence $v(t)>0$ for every $t>t_0$, and the
normalized update is well defined on $(t_0,t_1]$. Moreover, on every compact
subinterval $J\subset(t_0,t_1]$, the ratio $v(t)/g(t)^2$ is bounded away
from zero.

\medskip\noindent
\emph{Step 2: scalar decompositions and normalization.}
By Proposition~\ref{prop:m-v-delta-ap}, for all $t\in I$,
\[
m(t)
=
g(t)\bigl(1-\tau_1\delta(t)\bigr)
+
T_m(t)
+
C_m(t),
\]
and
\[
v(t)
=
g(t)^2\bigl(1-2\tau_2\delta(t)\bigr)
+
T_v(t)
+
C_v(t).
\]
Define
\[
q_m(t)
:=
\frac{T_m(t)+C_m(t)}{g(t)},
\qquad
q_v(t)
:=
\frac{T_v(t)+C_v(t)}{g(t)^2}.
\]
These quantities are well defined because $|g(t)|$ is bounded away from
zero on $I$. With this notation,
\[
m(t)
=
g(t)
\Bigl(
1-\tau_1\delta(t)+q_m(t)
\Bigr),
\]
and
\[
v(t)
=
g(t)^2
\Bigl(
1-2\tau_2\delta(t)+q_v(t)
\Bigr).
\]
Therefore, for $t>t_0$,
\[
R(t)
=
\frac{m(t)}{\sqrt{v(t)}}
=
\operatorname{sign}\bigl(g(t)\bigr)
\frac{
1-\tau_1\delta(t)+q_m(t)
}{
\sqrt{
1-2\tau_2\delta(t)+q_v(t)
}
}.
\]

\medskip\noindent
\emph{Step 3: linear and nonlinear separation of the ratio.}
Let
\[
x(t):=-\tau_1\delta(t)+q_m(t),
\qquad
y(t):=-2\tau_2\delta(t)+q_v(t).
\]
Then
\[
R(t)
=
\operatorname{sign}\bigl(g(t)\bigr)
F(x(t),y(t)),
\qquad
F(x,y):=\frac{1+x}{\sqrt{1+y}}.
\]
On any compact subinterval $J\subset(t_0,t_1]$, Step 1 implies that
$1+y(t)=v(t)/g(t)^2$ is bounded away from zero. Hence $F$ is smooth on
a neighbourhood of the curve $(x(t),y(t))$, $t\in J$.

Taylor's formula at $(0,0)$ gives
\[
F(x,y)
=
1+x-\frac12 y+N(x,y),
\]
where, on $J$,
\[
|N(x(t),y(t))|
\le
C_{R,J}\bigl(|x(t)|+|y(t)|\bigr)^2
\]
for some constant $C_{R,J}>0$. Substituting the definitions of $x(t)$ and
$y(t)$, the linear part becomes
\[
x(t)-\frac12 y(t)
=
-\tau_1\delta(t)+q_m(t)
+
\tau_2\delta(t)
-
\frac12 q_v(t).
\]
Therefore,
\[
x(t)-\frac12 y(t)
=
(\tau_2-\tau_1)\delta(t)
+
q_m(t)
-
\frac12 q_v(t).
\]
It follows that
\[
F(x(t),y(t))
=
1+(\tau_2-\tau_1)\delta(t)
+
q_m(t)
-
\frac12 q_v(t)
+
N_R(t),
\]
where $N_R(t):=N(x(t),y(t))$.

Consequently,
\[
R(t)
=
\operatorname{sign}\bigl(g(t)\bigr)
\Bigl(
1+(\tau_2-\tau_1)\delta(t)
\Bigr)
+
E_R(t),
\]
with
\[
E_R(t)
=
\operatorname{sign}\bigl(g(t)\bigr)
\left(
q_m(t)
-
\frac12 q_v(t)
+
N_R(t)
\right).
\]

Finally, since
\[
|x(t)|+|y(t)|
\le
3\tau_*|\delta(t)|+|q_m(t)|+|q_v(t)|,
\]
after increasing the constant if necessary we obtain, on $J$,
\[
|N_R(t)|
\le
C_{R,J}
\left(
\tau_*|\delta(t)|
+
|q_m(t)|
+
|q_v(t)|
\right)^2.
\]
This gives the announced bound for $E_R(t)$. Restoring the coordinate index
$i$ completes the proof.
\end{proof}

Theorem~\ref{thm:R-delta-ap} provides the rigorous version of the decomposition
used in the main text. The term
\[
(\tau_2-\tau_1)\operatorname{sign}\bigl(g_i(t)\bigr)\delta_i(t)
\]
is the explicit magnitude-lag contribution in the normalized update, and it
vanishes when the two Adam time scales coincide. The remaining term
$E_{R,i}(t)$ contains a linear contribution from the transition and curvature
terms inherited from $\mathbf{m}(t)$ and $\mathbf{v}(t)$, together with
the nonlinear effects introduced by the ratio
$\mathbf{m}(t)/\sqrt{\mathbf{v}(t)}$. In the empirical diagnostics, these
contributions are measured directly rather than treated as a negligible
asymptotic remainder.

\medskip\noindent
In what follows, briefly comments on how the above analysis extends to the more
standard practical variant of Adam that uses bias correction and a
nonzero stabilisation constant~$\varepsilon>0$ are provided.

\begin{remark}[Bias correction]
\label{rem:bias-correction}
In discrete time, Adam is often implemented with bias-corrected
moments,
\[
\hat{\mathbf{m}}_{k+1}
= \frac{\mathbf{m}_{k+1}}{1 - \beta_{1}^{k+1}},
\qquad
\hat{\mathbf{v}}_{k+1}
= \frac{\mathbf{v}_{k+1}}{1 - \beta_{2}^{k+1}},
\]
and the update
\[
\boldsymbol{\theta}_{k+1}
= \boldsymbol{\theta}_{k}
- \eta\,\frac{\hat{\mathbf{m}}_{k+1}}
{\sqrt{\hat{\mathbf{v}}_{k+1}} + \varepsilon}.
\]
Under the continuous-time scaling
\[
\beta_i = \exp(-h/\tau_i),
\qquad
t-t_0 = (k+1)h,
\]
the correction factors converge to
\[
a_i(t)^{-1}
=
\bigl(1-\exp(-(t-t_0)/\tau_i)\bigr)^{-1},
\qquad
a_i(t):=1-\exp(-(t-t_0)/\tau_i).
\]
Thus, for \(t>t_0\),
\[
\hat{\mathbf{m}}(t)
=
\frac{\mathbf{m}(t)}{a_1(t)},
\qquad
\hat{\mathbf{v}}(t)
=
\frac{\mathbf{v}(t)}{a_2(t)}.
\]

If the uncorrected moments satisfy
\[
\tau_1 \dot{\mathbf{m}}(t)
=
-\mathbf{m}(t)+\mathbf{g}(t),
\qquad
\tau_2 \dot{\mathbf{v}}(t)
=
-\mathbf{v}(t)+\mathbf{g}(t)^{\odot 2},
\]
then the bias-corrected variables satisfy the non-autonomous equations
\[
\tau_1 a_1(t)\dot{\hat{\mathbf{m}}}(t)
=
-\hat{\mathbf{m}}(t)+\mathbf{g}(t),
\qquad
\tau_2 a_2(t)\dot{\hat{\mathbf{v}}}(t)
=
-\hat{\mathbf{v}}(t)+\mathbf{g}(t)^{\odot 2}.
\]
Consequently, the exact continuous-time description of bias-corrected
Adam is not the same autonomous flow as in
\eqref{eq:adam-flow-app} with different initial data. Rather, it is a
non-autonomous flow whose effective relaxation times are
\[
\tau_i a_i(t)
=
\tau_i\bigl(1-\exp(-(t-t_0)/\tau_i)\bigr).
\]
These effective relaxation times vary on the initial layer
\(t-t_0=O(\tau_i)\) and converge exponentially fast to \(\tau_i\) as
\(t\to\infty\).

Thus bias correction should be viewed as an exponentially decaying
finite-time perturbation of the autonomous continuous-time Adam flow,
not merely as a change of initial condition. In particular, for
\(t-t_0\gg \tau_1,\tau_2\), the bias-corrected dynamics differs from
\eqref{eq:adam-flow-app} only by terms of size
\(O(\exp(-(t-t_0)/\tau_1))\) and
\(O(\exp(-(t-t_0)/\tau_2))\). Therefore, asymptotic statements that are
made after the initial transient, including the long-time structure of
the explicit magnitude-lag term involving
\((\tau_2-\tau_1)\delta_i(t)\), are unchanged up to exponentially
decaying remainder terms. However, finite-time statements near the
initial layer must use the non-autonomous equations above.
\end{remark}
\medskip\noindent
\begin{remark}[Nonzero $\varepsilon$]
\label{rem:epsilon}
If we keep $\varepsilon>0$ in the denominator, the normalized update in
continuous time becomes
\[
R_{\varepsilon,i}(t)
:=
\frac{m_i(t)}{\sqrt{v_i(t)}+\varepsilon}
\]
for each coordinate $i$. Equivalently,
\[
R_{\varepsilon,i}(t)
=
\frac{m_i(t)}{\sqrt{v_i(t)}}
\Bigl(1+\frac{\varepsilon}{\sqrt{v_i(t)}}\Bigr)^{-1}
=
R_i(t)
\Bigl(1+\frac{\varepsilon}{\sqrt{v_i(t)}}\Bigr)^{-1},
\]
where
\[
R_i(t):=\frac{m_i(t)}{\sqrt{v_i(t)}}
\]
denotes the corresponding update with $\varepsilon=0$.

Thus
\[
R_{\varepsilon,i}(t)
=
R_i(t)+\kappa_i(t),
\qquad
\kappa_i(t)
:=
-R_i(t)\,
\frac{\varepsilon}{\sqrt{v_i(t)}+\varepsilon}.
\]
Under the hypotheses of Theorem~\ref{thm:R-delta}, the decompositions for
$m_i(t)$ and $v_i(t)$ in Proposition~\ref{prop:m-v-delta} remain valid,
and $\sqrt{v_i(t)}$ is uniformly comparable to $|g_i(t)|$ on the
trajectory segment under consideration. Hence, whenever
$\varepsilon/|g_i(t)|$ is small, one has
\[
|\kappa_i(t)|
\lesssim
\frac{\varepsilon}{|g_i(t)|},
\]
with constants depending only on the same uniform bounds as in
Theorem~\ref{thm:R-delta}. Combining this with the decomposition in
Theorem~\ref{thm:R-delta} gives
\[
R_{\varepsilon,i}(t)
=
\operatorname{sign}\bigl(g_i(t)\bigr)
\Bigl(1+(\tau_2-\tau_1)\delta_i(t)\Bigr)
+
E_{R,i}(t)
+
\kappa_i(t),
\]
where $E_{R,i}(t)$ is the remainder already present in
Theorem~\ref{thm:R-delta}, while $\kappa_i(t)$ is the additional
stabilization error induced by $\varepsilon$.

The term $\kappa_i(t)$ has a distinct dependence on
$\varepsilon/|g_i(t)|$ and is therefore retained separately from
$E_{R,i}(t)$. Its effect is already visible for a constant nonzero gradient. For fixed
$\varepsilon>0$, the quantity $\varepsilon/|g_i(t)|$ is independent of
the magnitude-drift bound $\Lambda$. Therefore, in the asymptotic regime
$\Lambda\to0$, the term $\kappa_i(t)$ need not vanish and cannot be
absorbed into the tracking error. This obstruction is already visible in the idealized case of a
perfectly flat gradient trajectory. If $g_i(t)\equiv g_i\neq0$, then
$m_i(t)=g_i$ and $v_i(t)=g_i^2$ after transients, so that
\[
R_{\varepsilon,i}(t)
=
\frac{g_i}{|g_i|+\varepsilon}
=
\operatorname{sign}(g_i)
\Bigl(1+\frac{\varepsilon}{|g_i|}\Bigr)^{-1}.
\]
Thus, even when the magnitude-drift term vanishes, the update differs from
the sign baseline $\operatorname{sign}(g_i)$ by an amount of
order $\varepsilon/|g_i|$. Consequently, a nonzero stabilization
constant introduces an additional dependence on the absolute coordinate magnitude.

The zero-$\varepsilon$ decomposition of Theorem~\ref{thm:R-delta} therefore
remains a good approximation only in regimes where
$\varepsilon$ is negligible compared with the relevant coordinatewise
gradient magnitudes. If $\varepsilon$ is comparable to, or larger than,
$|g_i(t)|$, the denominator is significantly affected by the absolute
magnitude set by $\varepsilon$, and the idealized zero-\(\varepsilon\) flow no
longer accurately describes the algorithm in that coordinate.
\end{remark}

\section{Additional Details for the Literal Discrete Decomposition}
\label{appendix:practical_validity}

The decomposition in Theorem~\ref{thm:R-delta} is local in time and
coordinate. We instantiate its sign-stable regime on gradients generated by
ordinary training, without constructing a synthetic signal or selecting
coordinates according to the resulting decomposition error.

\subsection{Training trajectory and discrete term}

We use the Digits classification dataset with a stratified $80/20$
train--test split. Inputs are standardized using statistics fitted on the
training split. The model is a $64$--$32$--$32$--$10$ MLP with Tanh
activations and $3466$ trainable parameters. It is trained from scratch for
$1500$ full-batch steps using PyTorch Adam with learning rate $3\times10^{-4}$,
weight decay zero, and $\varepsilon=10^{-8}$. Every pair
$(\beta_1,\beta_2)\in\{0.9,0.99,0.999\}^2$ is trained separately for seeds
$0$, $1$, and $2$. The actual full training gradient is stored immediately
before each parameter update. Further implementation details are given in
Appendix~\ref{appendix:implementation_details}.

Let $u_{i,k}=\log|g_{i,k}|$. For each $\beta$, we compute
\begin{equation}
 h_{\beta,i,k}
 =\beta h_{\beta,i,k-1}+(1-\beta)u_{i,k},
 \qquad
 \bar u_{\beta,i,k}
 =\frac{h_{\beta,i,k}}{1-\beta^{k+1}},
 \label{eq:appendix-discrete-log-ema}
\end{equation}
with $h_{\beta,i,-1}=0$. Index $k=0$ denotes the gradient stored before the
first parameter update. The moment states are computed as
\begin{align}
 m_{\beta,i,k}
 &=\beta m_{\beta,i,k-1}+(1-\beta)g_{i,k},
 &M_{\beta,i,k}&=\frac{m_{\beta,i,k}}{1-\beta^{k+1}},\\
 v_{\beta,i,k}
 &=\beta v_{\beta,i,k-1}+(1-\beta)g_{i,k}^{2},
 &V_{\beta,i,k}&=\frac{v_{\beta,i,k}}{1-\beta^{k+1}},
\end{align}
with $m_{\beta,i,-1}=v_{\beta,i,-1}=0$, and
\begin{equation}
 R_{i,k}=\frac{M_{\beta_1,i,k}}
 {\sqrt{V_{\beta_2,i,k}}+10^{-8}}.
\end{equation}
The stored float32 gradient arrays are loaded in float64, and all subsequent
EMA calculations and metrics use float64. To keep the full log-EMA state
defined, the recursion uses
$\log\max\{|g_{i,k}|,\operatorname{tiny}_{64}\}$, where
$\operatorname{tiny}_{64}$ is the smallest positive normal float64 value.
The validity mask excludes every evaluation window containing an exact zero;
no additional magnitude threshold is applied. Non-finite evaluations are
also excluded.

To obtain the discrete first-order term, write a sign-stable gradient as
$g_{i,k}=S_{i,k}\exp(u_{i,k})$. Expanding the two bias-corrected memories to
first order in the variation of $u$ gives
\begin{equation}
 \frac{M_{\beta_1,i,k}}{\sqrt{V_{\beta_2,i,k}}}
 =S_{i,k}\left[1+\bar u_{\beta_1,i,k}
 -\bar u_{\beta_2,i,k}\right]+O(\|\Delta u\|^2).
\end{equation}
This yields the full-history term in
Eq.~\eqref{eq:literal-discrete-lag}. For a linear trace
$u_{i,k}=u_{i,0}+k\delta_i$, it satisfies
\begin{equation}
 \bar u_{\beta_1,i,k}-\bar u_{\beta_2,i,k}
 \longrightarrow
 (\ell_2-\ell_1)\delta_i,
 \qquad \ell_r=\frac{\beta_r}{1-\beta_r},
\end{equation}
recovering the explicit term in Theorem~\ref{thm:R-delta}. The measured
remainder $E=R-S-L^{\mathrm{disc}}$ retains transition, curvature, higher-order
ratio, discretization, and $\varepsilon$ effects.

\subsection{Aggregation and robustness}

A coordinate-time sample at step $k$ is valid when
$g_{i,k-w+1},\ldots,g_{i,k}$ are all nonzero and have the same sign. This
condition selects evaluation points but never resets an EMA state. The main
experiment uses the same $w=1000$ for every row. For
$X\in\{S,L^{\mathrm{disc}},E\}$, its absolute budget is
\begin{equation}
 B_X=100\,
 \frac{\sum_{(i,k)\in\mathcal D}|X_{i,k}|}
 {\sum_{(i,k)\in\mathcal D}
 (|S_{i,k}|+|L^{\mathrm{disc}}_{i,k}|+|E_{i,k}|)}.
\end{equation}
The explained fraction $F_2$ is defined in
Section~\ref{subsec:decomposition-diagnostics}. To diagnose cancellation, we
also compute
\begin{equation}
 \kappa=\frac{\sum|L^{\mathrm{disc}}|+\sum|E|}
 {\sum|L^{\mathrm{disc}}+E|},
\end{equation}
where $\kappa=1$ means no cancellation. For the conventional off-diagonal
configurations, $\kappa$ ranges from $1.04$ to $1.68$ with $w=1000$ and from
$1.25$ to $1.78$ with $w=6$. These values show that the decomposition does not
depend on near-complete cancellation between the lag and residual terms. Sign
agreement is computed over nonzero $L^{\mathrm{disc}}$ and $R-S$ as the
fraction for which their product is positive.

As a deliberately less restrictive control, Table~\ref{tab:literal-w6}
repeats the conventional off-diagonal rows with $w=6$. The residual remains
between $14\%$ and $16\%$, so the result in
Table~\ref{tab:literal-discrete-decomposition} is not created by the long
sign-stability window.

\begin{table}[h]
\centering
\small
\caption{Short-window robustness of the literal decomposition. Values are
means over three independently trained seeds.}
\label{tab:literal-w6}
\begin{tabular*}{\columnwidth}{@{\extracolsep{\fill}}lrrrrr@{}}
\toprule
$(\beta_1,\beta_2)$ & $S$ & $L^{\rm disc}$ & $E$ & $\kappa$ & $F_2$ \\
\midrule
$(0.9,0.99)$   & $60.60\%$ & $25.26\%$ & $14.15\%$ & $1.25$ & $59.2\%$ \\
$(0.9,0.999)$  & $43.67\%$ & $40.04\%$ & $16.29\%$ & $1.78$ & $38.5\%$ \\
$(0.99,0.999)$ & $49.94\%$ & $35.23\%$ & $14.82\%$ & $1.46$ & $67.7\%$ \\
\bottomrule
\end{tabular*}
\end{table}

We also evaluated the reverse ordering $\beta_1>\beta_2$ as a stress test,
using a sign window matched to the longer memory. The residual budgets are
$23.98\%$ for $(0.99,0.9)$ and $23.61\%$ for $(0.999,0.99)$. The extreme
pair $(0.999,0.9)$ has a larger $45.87\%$ residual and only approximately
$0.03$ million valid evaluations per seed. It also trains markedly worse,
reaching $88.7\%$ mean test accuracy, compared with $95.2$--$97.1\%$ across
the diagonal and conventional-order rows. We therefore treat this corner as
a non-perturbative stress case rather than evidence for the main quantitative
claim, which concerns the conventional Adam ordering $\beta_1\leq\beta_2$.

\section{Additional Data on Experiments}
\label{appendix:additional_experiments}

\subsection{Training Dynamics Across Tasks}
In this section, we report the training dynamics for all model--dataset pairs considered in our study. These plots follow the protocol described in Section~\ref{subsec:update-norm-dynamics}; shaded regions denote one standard deviation across seeds.

In all figures, we show the training dynamics for the corresponding
model and dataset. Each panel displays the evolution of the training loss
(left axis) and the norm of the update $\mathbf{R}_k$ (right axis) for a
fixed pair $(\beta_{1},\beta_{2})$. Rows correspond to fixed values of
$\beta_{1}$, while columns vary $\beta_{2}$, with identical axis scales
within each row to enable direct comparison. Curves represent the
seed-averaged dynamics after exponential smoothing with a window of 200
steps, and shaded regions indicate one standard deviation across random
seeds. The oscillation metric $\omega$, reported in the title of each
panel, is computed as the mean across seeds for the corresponding pair
$(\beta_{1},\beta_{2})$.

\begin{figure*}[ht]
    \centering
        \includegraphics[width=\textwidth]{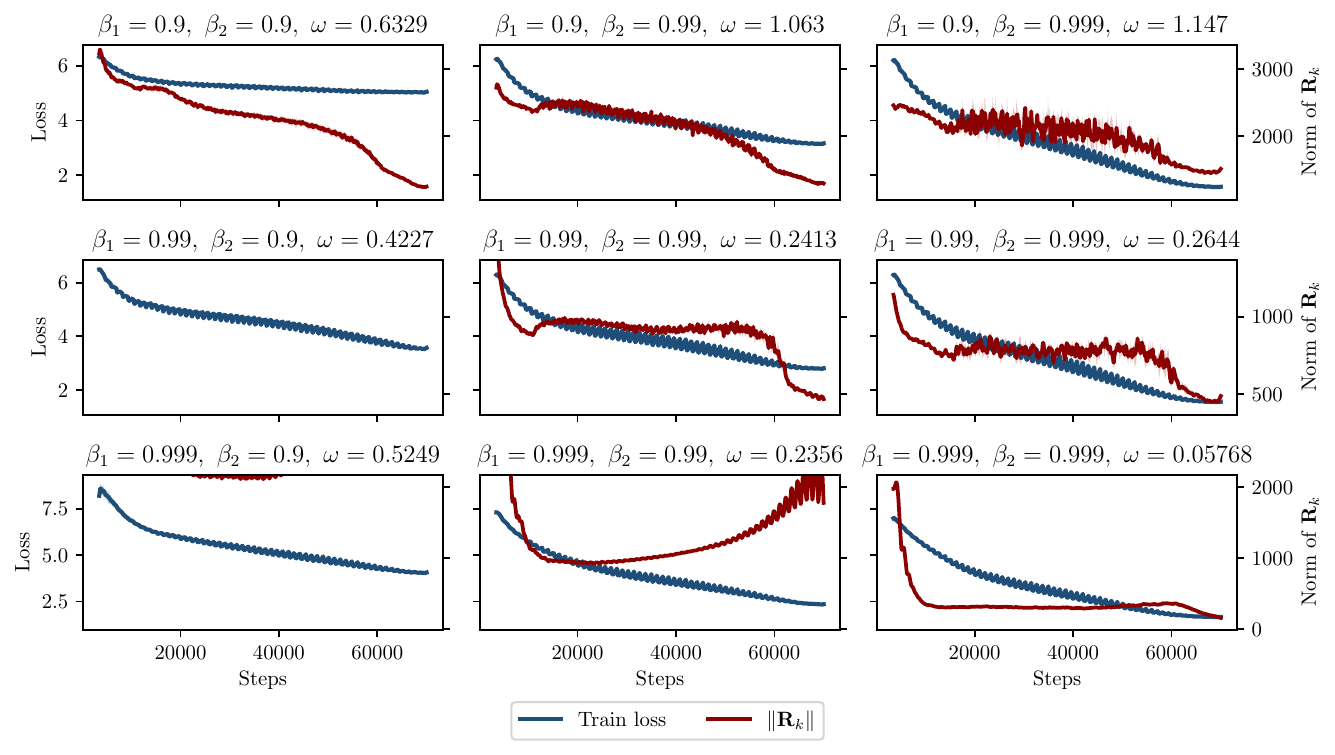}
\caption{%
Training dynamics for NanoGPT on WikiText.}

\label{fig:betas-1}
\end{figure*}

Across all configurations, the qualitative behavior observed in these
figures closely mirrors that reported in the main text. In particular,
when comparing panels row-wise, configurations along the diagonal
$\beta_{1}=\beta_{2}$ typically exhibit smoother update norms and
reduced oscillatory behavior, while off-diagonal choices lead to visibly
larger fluctuations. This visual consistency across architectures and datasets shows that the same qualitative association recurs across all six tasks.

\begin{figure*}[!ht]
    \centering
    \includegraphics[width=\textwidth]{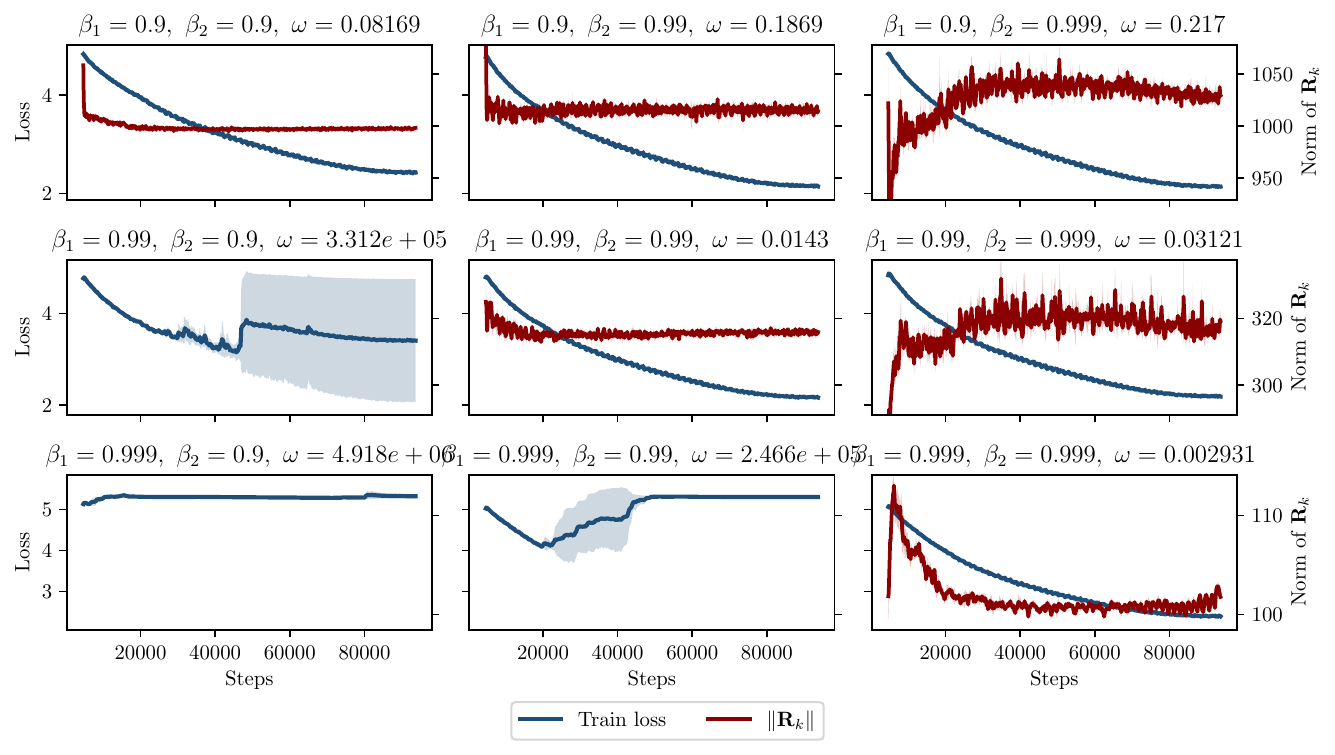}
    \caption{Training dynamics for EfficientNet-B0 on TinyImageNet.}
    \label{fig:betas-efficientnet-tinyimagenet}
\end{figure*}

\begin{figure*}[!ht]
    \centering
    \includegraphics[width=\textwidth]{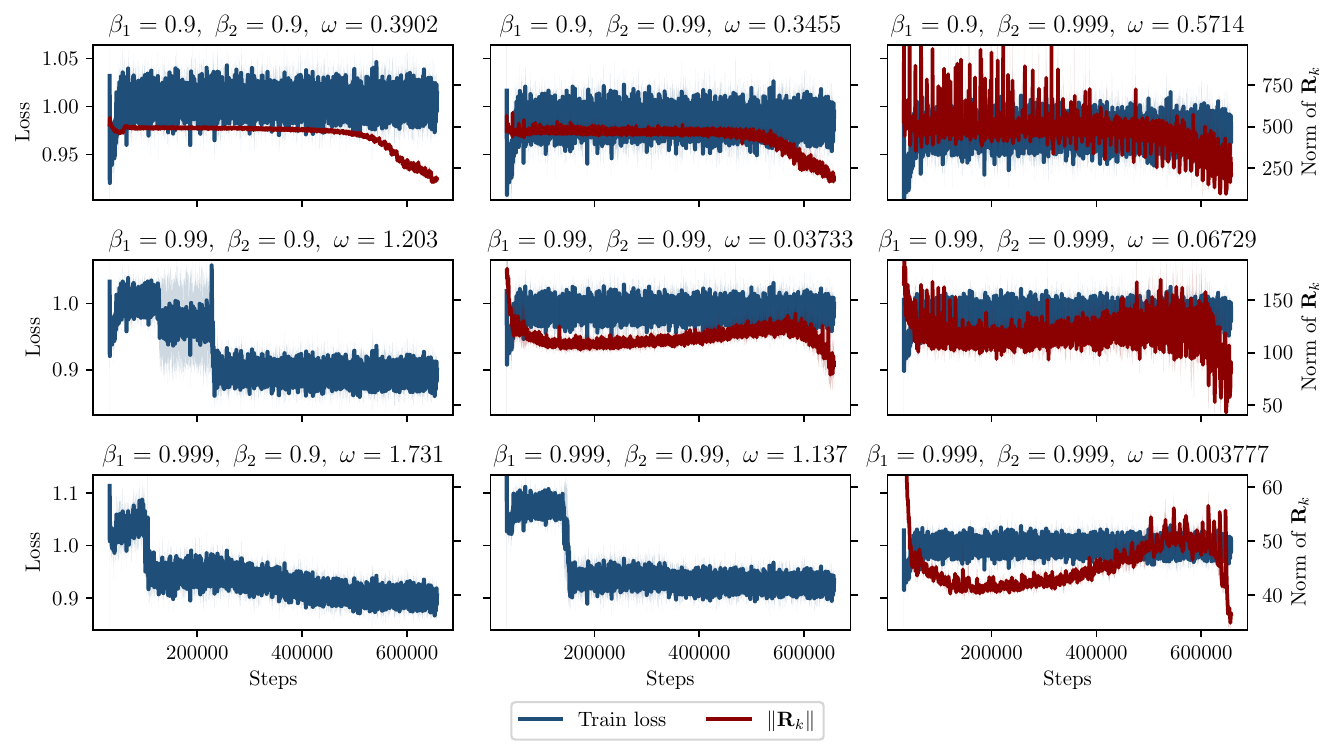}
    \caption{Training dynamics for T5 on SQuAD.}
    \label{fig:betas-t5-squad}
\end{figure*}

\begin{figure*}[!ht]
    \centering
    \includegraphics[width=\textwidth]{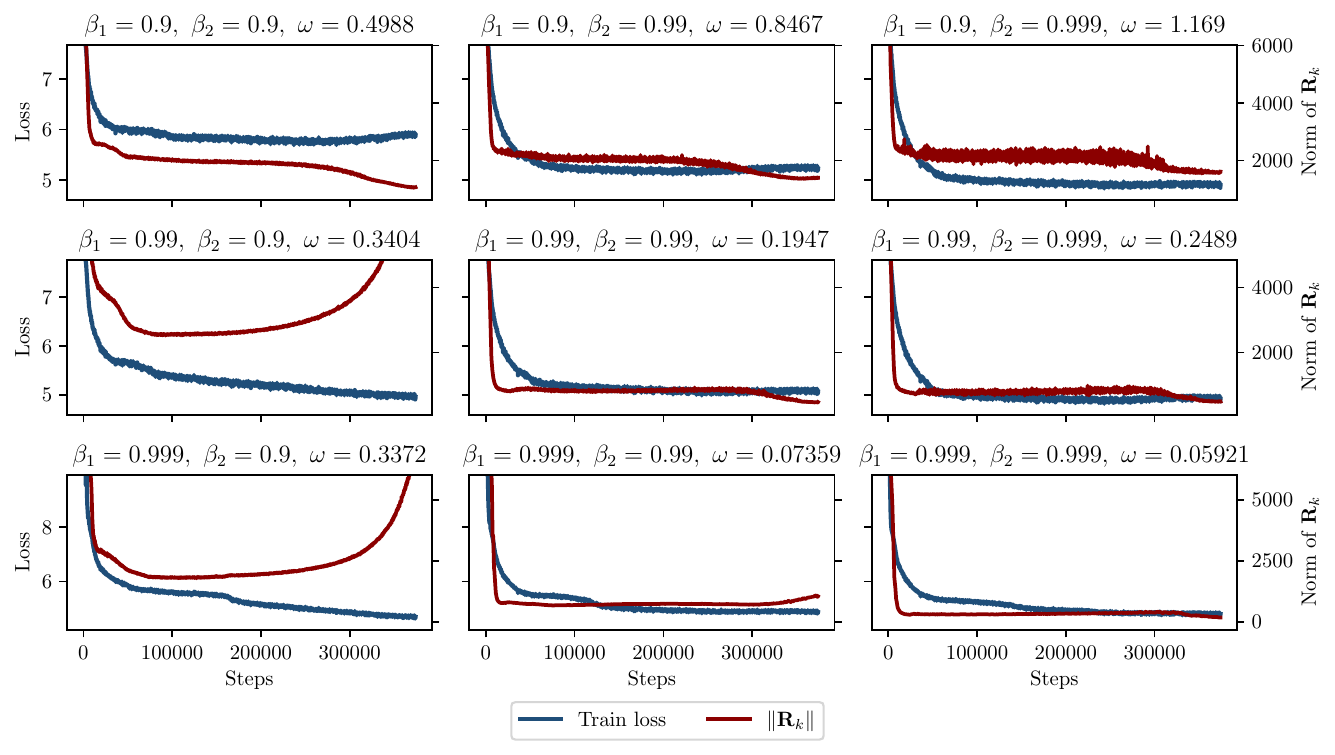}
    \caption{Training dynamics for NanoGPT on SlimPajama.}
    \label{fig:betas-nanogpt-slimpajama}
\end{figure*}

\begin{figure*}[!ht]
    \centering
    \includegraphics[width=\textwidth]{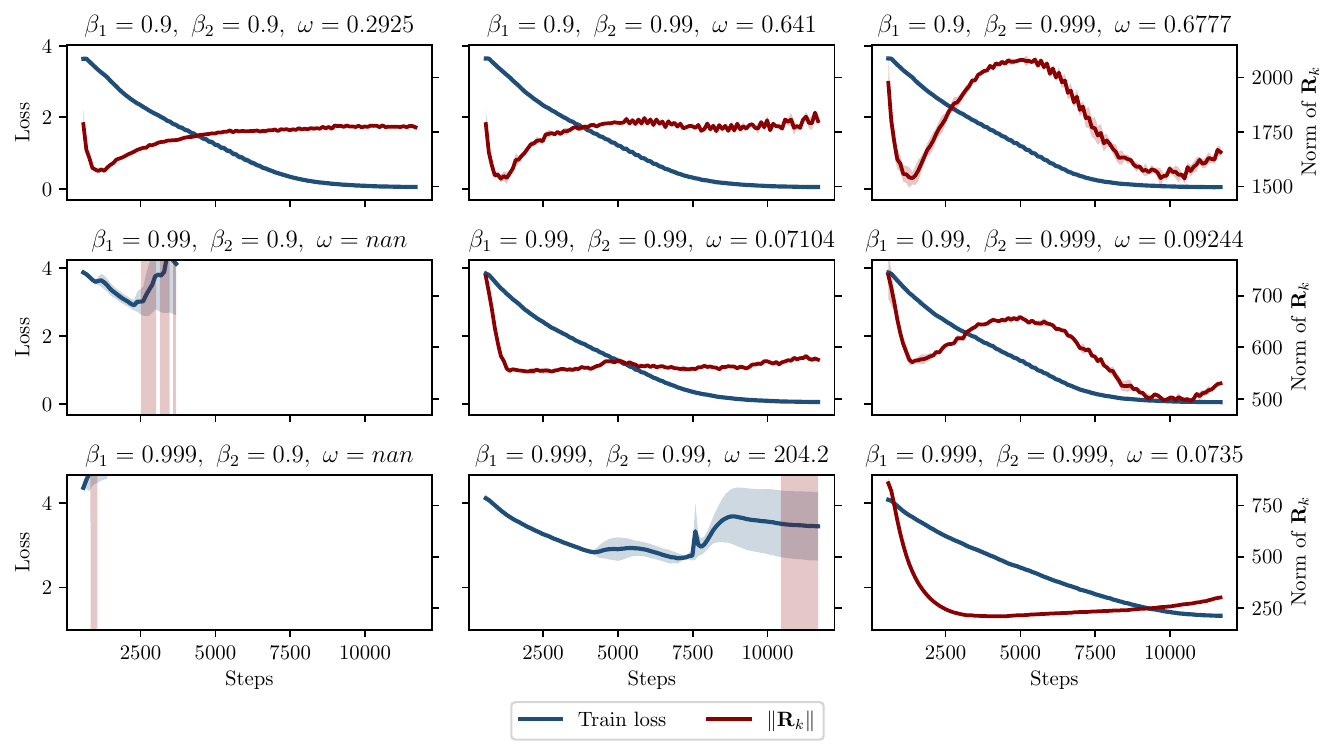}
    \caption{Training dynamics for ViT-B16 on CIFAR-100.}
    \label{fig:betas-vit-cifar100}
\end{figure*}

\begin{figure*}[!ht]
    \centering
    \includegraphics[width=0.9\textwidth]{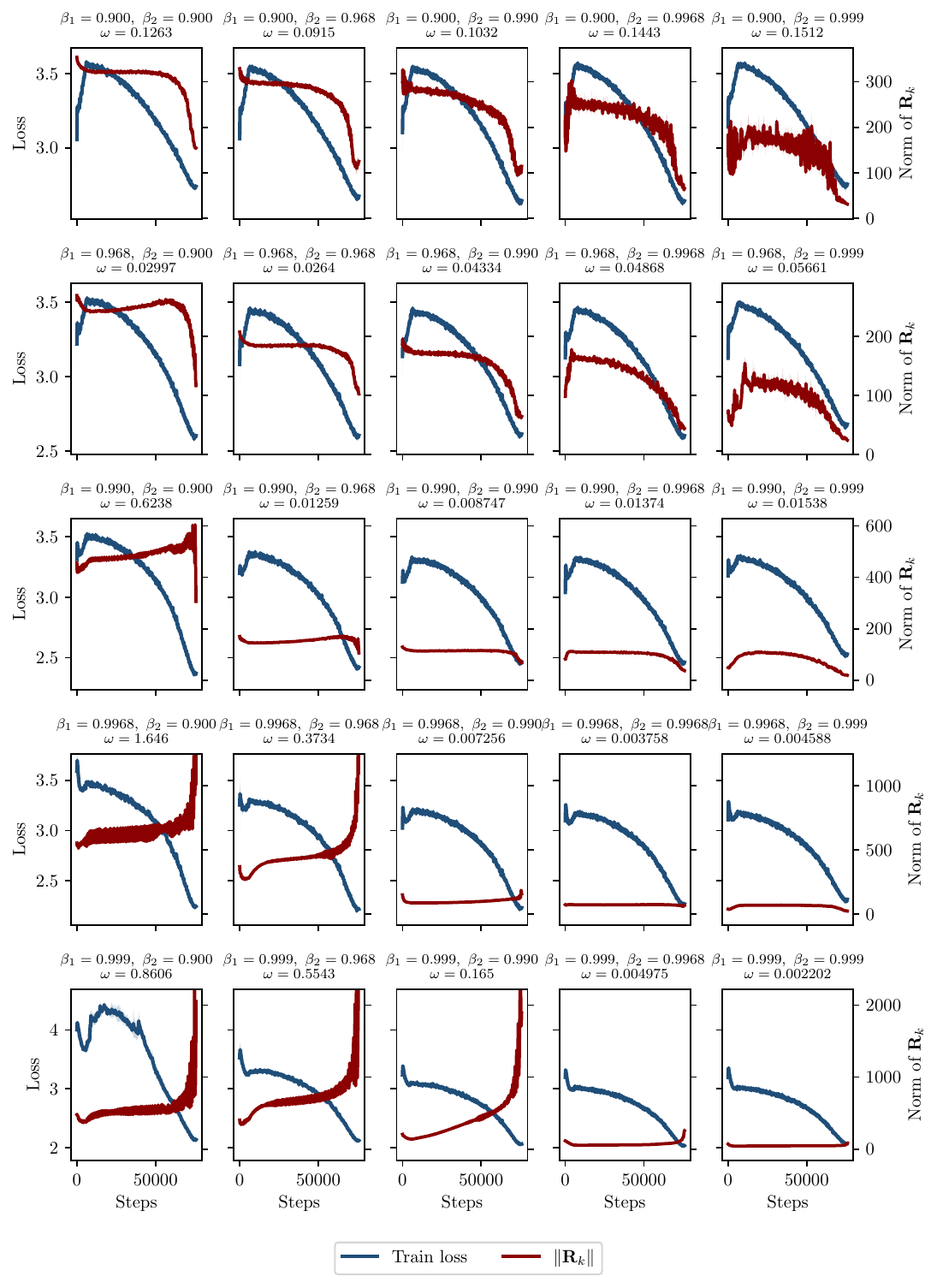}
    \caption{Training dynamics for ResNet18 on CIFAR-100.}
    \label{fig:betas-resnet-cifar100}
\end{figure*}

\clearpage

\subsection{Ablation on the Oscillation Metric and Smoothing Window}

We further assess the robustness of our conclusions with respect to both
the definition of the oscillation metric and the choice of the smoothing
window. In the main experiments, oscillation is quantified using a
first-difference metric, denoted by $\omega_{1}$ ($\omega$ in the main text), which measures the average
magnitude of consecutive variations in the update norm. Given the smoothed sequence $\{x_k\}_{k=1}^{T}$ derived from an EMA of the sequence $\|\mathbf{R}_k\|$, this
metric is defined as
\[
\omega_{1}(x)
\;=\;
\frac{1}{T-1}\sum_{k=1}^{T-1}\bigl|x_{k+1}-x_k\bigr|,
\]
capturing the typical step-to-step variation of the update magnitude.

In addition, we consider an alternative second-difference metric,
denoted by $\omega_{2}$, based on discrete second differences. This
metric measures the overall magnitude of curvature in the sequence
$\{x_k\}$ and is sensitive to higher-frequency fluctuations. Concretely,
we set
\[
\omega_{2}(x)
=
\frac{1}{T-2}
\sum_{k=2}^{T-1}
\left|x_{k+1}-2x_k+x_{k-1}\right|,
\]
with suitable one-sided differences at the boundaries. While $\omega_{1}$ directly reflects local variability, $\omega_{2}$
captures rapid changes in slope and emphasizes oscillatory behavior at
shorter time scales.

For both metrics, we vary the exponential smoothing window over the set
$\{1,10,100,200,500\}$. For brevity, we do not
report the raw oscillation values; instead, we summarize the results in
terms of the selection rate of the diagonal $\beta_{1}=\beta_{2}$ reported in
Table~\ref{tab:ablation_windows_metrics}.

\begin{table*}[ht]
\centering
\small
\renewcommand{\arraystretch}{1.15}
\caption{%
Ablation results for the smoothing window and the oscillation metric.
For each model--dataset pair and window size, we report the rate at which
the diagonal $\beta_{1}=\beta_{2}$ minimizes the oscillation.
Results are shown for the first-difference metric $\omega_1$, based on
absolute first differences of the smoothed $\|\mathbf{R}_k\|$, and for
the second-difference metric $\omega_2$, based on discrete second differences.}
\label{tab:ablation_windows_metrics}

\begin{minipage}[t]{0.49\textwidth}
\centering
\begin{tabular}{@{}lccc@{}}
\toprule
Model / Dataset &
Window &
$\omega_1$ Rate &
$\omega_2$ Rate \\
\midrule

\multirow[c]{5}{*}{\shortstack[l]{NanoGPT\\SlimPajama}}
& 1   & 100\% & 100\% \\
& 10  & 67\%  & 100\% \\
& 100 & 100\% & 100\% \\
& 200 & 100\% & 100\% \\
& 500 & 100\% & 100\% \\
\midrule

\multirow[c]{5}{*}{\shortstack[l]{EfficientNet-B0\\TinyImageNet}}
& 1   & 100\% & 100\% \\
& 10  & 100\% & 100\% \\
& 100 & 100\% & 100\% \\
& 200 & 100\% & 100\% \\
& 500 & 100\% & 100\% \\
\midrule

\multirow[c]{5}{*}{\shortstack[l]{T5\\SQuAD}}
& 1   & 67\% & 56\% \\
& 10  & 67\% & 67\% \\
& 100 & 67\% & 67\% \\
& 200 & 78\% & 67\% \\
& 500 & 78\% & 67\% \\

\bottomrule
\end{tabular}
\end{minipage}
\hfill
\begin{minipage}[t]{0.49\textwidth}
\centering
\begin{tabular}{@{}lccc@{}}
\toprule
Model / Dataset &
Window &
$\omega_1$ Rate &
$\omega_2$ Rate \\
\midrule

\multirow[c]{5}{*}{\shortstack[l]{ResNet18\\CIFAR-100}}
& 1   & 33\% & 7\% \\
& 10  & 40\% & 0\% \\
& 100 & 80\% & 13\% \\
& 200 & 80\% & 33\% \\
& 500 & 80\% & 33\% \\
\midrule

\multirow[c]{5}{*}{\shortstack[l]{NanoGPT\\WikiText}}
& 1   & 56\%  & 56\% \\
& 10  & 44\%  & 67\% \\
& 100 & 89\%  & 44\% \\
& 200 & 100\% & 56\% \\
& 500 & 100\% & 56\% \\
\midrule

\multirow[c]{5}{*}{\shortstack[l]{ViT-B16\\CIFAR-100}}
& 1   & 100\% & 100\% \\
& 10  & 100\% & 100\% \\
& 100 & 100\% & 100\% \\
& 200 & 100\% & 100\% \\
& 500 & 100\% & 100\% \\

\bottomrule
\end{tabular}
\end{minipage}

\end{table*}

Focusing first on $\omega_{1}$, the metric used throughout the main article, we observe that the results are stable for sufficiently large windows. For smaller windows, stochastic minibatch fluctuations partly obscure the oscillation signal, leading to less consistent diagonal selection despite similar overall trends.

For $\omega_1$, diagonal selection becomes more frequent once the stochastic
variation is attenuated by moderate smoothing. The results for $\omega_2$ are
more heterogeneous, indicating that first and second differences capture
different aspects of the update-norm trajectory.

\newpage
\section{Implementation Details}\label{appendix:implementation_details}

\paragraph{Common setup.}
All experiments evaluated Adam-type optimizers over a $3\times3$ momentum grid 
$(\beta_1,\beta_2)\in\{0.9,0.99,0.999\}^2$,
except for ResNet18 on CIFAR-100, where we use a finer $5\times5$ grid
$\beta_{1},\beta_{2}\in\{0.9,0.968,0.99,0.9968,0.999\}$.
Models were initialized using the same sinusoidal scheme \citep{sinusoidal} for accelerating training and trained in full FP32 precision, disabling both mixed precision and TF32 operations.
Whenever specified, random seeds were fixed for NumPy and PyTorch.
Learning rates followed a linear warm-up phase up to the base value, followed by cosine annealing for the remainder of training using \texttt{SequentialLR}.

\paragraph{Experiment 1 (NanoGPT, SlimPajama).}
Training employed Adam with base learning rate $3\times10^{-4}$ and weight decay $10^{-4}$ for one epoch.
Warm-up covered $20\%$ of the total training steps, starting from a factor $10^{-2}$ of the base learning rate.
Gradient clipping with maximum norm $1$ was enabled.
The model followed a GPT-2 small configuration with 12 layers, 12 attention heads, hidden size 768, block size 1024, dropout $0.1$, and bias terms enabled.
Training and validation used batch size $8$ and sequence length $1024$.
Text was tokenized with the GPT-2 tokenizer through \texttt{tiktoken}, and the language model was trained from scratch.

\paragraph{Experiment 2 (EfficientNet-B0, TinyImageNet).}
Optimization relied on AdamW with learning rate $10^{-3}$ and weight decay $3\times10^{-5}$ over 60 epochs.
Warm-up spanned $20\%$ of the training epochs, starting from a factor $10^{-1}$.
The classifier head was replaced by a dropout layer with rate $0.6$ followed by a linear projection to 200 classes.
Training used cross-entropy loss with label smoothing $0.1$ and batch size $64$.
Images were processed at resolution $224\times224$.
Training augmentation used random resized cropping, horizontal and vertical flips, rotations up to $15^\circ$, and \texttt{ColorJitter} with parameters $0.4/0.4/0.4/0.1$; inputs were normalized using ImageNet statistics.

\paragraph{Experiment 3 (T5, SQuAD).}
Adam was used with learning rate $10^{-3}$ and weight decay $10^{-2}$ for 60 epochs.
Warm-up lasted $20\%$ of epochs, starting from a factor $10^{-2}$.
Training minimized the model-provided cross-entropy loss returned by the forward pass.
Training and validation used batch sizes $8$ and $16$, respectively, with the \texttt{t5-small} tokenizer.
Inputs were formatted as \texttt{question: <question> context: <context>}, with the first annotated answer used as target.
Input and target sequences were padded and truncated to maximum lengths $256$ and $64$, respectively, and evaluation used the SQuAD validation split.

\paragraph{Experiment 4 (ResNet18, CIFAR-100).}
Training employed Adam with learning rate $2\times10^{-2}$ and weight decay $10^{-3}$ over 100 epochs while sweeping all $(\beta_1,\beta_2)$ combinations.
Warm-up was applied over the first 10 epochs, starting from a factor $0.1$ of the base learning rate.
Standard cross-entropy loss and batch size $64$ were used.
Training images underwent random horizontal flipping and random $32\times32$ cropping with padding $4$, followed by normalization using ImageNet statistics.
Evaluation used the official CIFAR-100 test split without training-time augmentation.

\paragraph{Experiment 5 (NanoGPT, WikiText-2).}
Adam with learning rate $3\times10^{-4}$ and weight decay $10^{-4}$ was used for 60 epochs.
Warm-up covered $20\%$ of training steps, starting from a factor $10^{-2}$.
Gradient clipping with norm $1$ was enabled.
The architecture matched GPT-2 small (12 layers, 12 heads, hidden size 768, block size 1024, dropout $0.1$).
Training and validation used batch sizes $8$ and $16$, respectively.
WikiText-2 was tokenized with the GPT-2 tokenizer; empty text entries were removed, the remaining text was concatenated, and the resulting token stream was partitioned into blocks of length $256$.
Inputs and targets were obtained by a one-token shift within each block, and evaluation used the validation split.

\paragraph{Experiment 6 (ViT-B16, CIFAR-100).}
Optimization employed AdamW with learning rate $3\times10^{-4}$ and weight decay $0.2$ over 60 epochs.
Warm-up spanned $10\%$ of epochs, starting from a factor $0.1$.
Cross-entropy loss was used with batch size $256$.
Training preprocessing consisted of a $224\times224$ random resized crop with scale range $0.8$--$1.0$, bicubic interpolation, random horizontal flipping, color jitter with strength $0.2$, and random erasing with probability $0.1$.
Inputs were normalized using CIFAR-100 statistics; validation images were resized to $256\times256$ and center-cropped to $224\times224$.
Non-finite losses triggered skipped updates, and non-finite gradients and optimizer states were sanitized via \texttt{nan\_to\_num}.

\paragraph{Hardware.}
All experiments were executed on a multi-GPU server equipped with NVIDIA A100-SXM4 GPUs with 80\,GB of HBM2 memory each. 
Training runs typically utilized up to three GPUs concurrently, each operating at full utilization during active workloads. 
The host system featured dual AMD EPYC 7513 processors (64 physical cores in total) with a NUMA architecture across eight nodes. 
All computations were performed under x86\_64 Linux in full FP32 precision.

\paragraph{Synthetic magnitude-variation experiment.}
The synthetic experiment in Figures~\ref{fig:scale-experiment1} and
\ref{fig:scale-experiment2} uses the deterministic gradient sequence
\[
\mathbf{g}_k = s_k\,\frac{(10,\,-8+\sin(k/10),\,9+\cos(k/10))}
{\|(10,\,-8+\sin(k/10),\,9+\cos(k/10))\|},
\qquad k=0,\ldots,600,
\]
where
\[
s_k
=
1+\frac12\left[
\sigma\!\left(\frac{k-200}{20}\right)
-
\sigma\!\left(\frac{k-400}{20}\right)
\right],
\qquad
\sigma(x)=\frac{1}{1+e^{-x}}.
\]
Thus the gradient norm undergoes a smooth magnitude increase centered at
step \(200\), a plateau, and a smooth decrease centered at step \(400\).
For each tested pair \((\beta_1,\beta_2)\), Adam moments are initialized
with \(m_0=v_0=0\) and updated with bias correction, learning rate
\(10^{-3}\), and \(\varepsilon=10^{-8}\). The plotted quantity is
\[
\mathbf{R}_k
=
\frac{\hat{\mathbf{m}}_k}{\sqrt{\hat{\mathbf{v}}_k}+\varepsilon},
\]
shown through its Euclidean norm. Figure~\ref{fig:scale-experiment1} uses
\(\beta_1=\beta_2\in\{0.9,0.95,0.99,0.999\}\), while
Figure~\ref{fig:scale-experiment2} uses
\[
(\beta_1,\beta_2)\in
\{(0.9,0.999),(0.9,0.95),(0.9,0.99),(0.99,0.999)\}.
\]
No randomness is used in this experiment.

\paragraph{Literal discrete decomposition.}
The experiment of Table~\ref{tab:literal-discrete-decomposition} uses the
scikit-learn Digits dataset, split into $1437$ training and $360$ test
examples by a stratified $80/20$ split whose random state equals the run
seed. Inputs are standardized coordinatewise using only training-set
statistics. The $64$--$32$--$32$--$10$ Tanh MLP is trained from scratch for
$1500$ full-batch steps. We use PyTorch Adam with learning rate
$3\times10^{-4}$, weight decay zero, and $\varepsilon=10^{-8}$, placed
outside the square root as in Eq.~\eqref{eq:adam-theta}. Every pair in
$\{0.9,0.99,0.999\}^2$ is trained independently for seeds $0$, $1$, and $2$
with deterministic PyTorch algorithms enabled. Immediately before each
optimizer update, the complete gradient vector is stored in float32 in
PyTorch parameter order, giving an array of shape $1500\times3466$ for each
run. The analysis loads the full array in float64 and uses every coordinate.
The moment and log-magnitude EMAs start from zero at step $k=-1$, retain their
full history, and use bias correction $1-\beta^{k+1}$. Evaluation discards
the first $200$ steps and retains endpoint $k$ only when the preceding $w$
gradients, including both endpoints, are all nonzero and have identical
sign. The main table uses $w=1000$; Table~\ref{tab:literal-w6} reports the
$w=6$ control. Exact-zero and non-finite samples are excluded, but no
magnitude threshold or coordinate subsampling is used. The normalized update
is $R=M/(\sqrt{V}+10^{-8})$, and all reported statistics are means over the
three independently trained seeds.

\end{document}